\documentclass{article}

\usepackage[utf8]{inputenc} %
\usepackage[T1]{fontenc}    %

\usepackage[accepted]{icml2024}

\usepackage{natbib}
\usepackage{hyperref}       %
\usepackage{url}            %
\usepackage{booktabs}       %
\usepackage{amsfonts}       %
\usepackage{nicefrac}       %
\usepackage{mathtools}
\usepackage[algo2e]{algorithm2e} 
\usepackage{nicefrac}
\usepackage[makeroom]{cancel}
\usepackage{listings}
\usepackage{enumitem}

\usepackage{amsmath}
\usepackage{amssymb}
\usepackage{mathtools}
\usepackage{amsthm}

\usepackage{float}
\usepackage{tikz}
\usepackage{caption}

\definecolor{codegreen}{rgb}{0,0.6,0}
\definecolor{codegray}{rgb}{0.5,0.5,0.5}
\definecolor{codepurple}{rgb}{0.58,0,0.82}
\definecolor{backcolour}{rgb}{0.95,0.95,0.92}

\definecolor{rgreen}{rgb}{0,0.4,0.1}
\definecolor{deepblue}{rgb}{0,0.023,0.694}

\lstdefinestyle{mystyle}{
    backgroundcolor=\color{backcolour},   
    commentstyle=\color{codegreen},
    keywordstyle=\color{magenta},
    numberstyle=\tiny\color{codegray},
    stringstyle=\color{codepurple},
    basicstyle=\ttfamily\scriptsize,
    breakatwhitespace=false,         
    breaklines=true,                 
    captionpos=b,                    
    keepspaces=true,                 
    numbers=left,                    
    numbersep=5pt,                  
    showspaces=false,                
    showstringspaces=false,
    showtabs=false,                  
    tabsize=2
}
\newfloat{lstfloat}{htbp}{lop}
\floatname{lstfloat}{Listing}

\newcommand{\E}[2]{\mathop{\mathlarger{\mathbb{E}} }_{#1} #2}

\newcommand{\calD}{\mathcal D}

\newcommand{\calM}{\mathcal M}

\newcommand{\calO}{\mathcal O}

\newcommand{\calY}{\mathcal Y}

\newcommand{\calL}{\mathcal L}

\newcommand{\prm}{\theta}
\newcommand{\prmhat}{\hat{\theta}}
\newcommand{\prmmle}{\hat{\theta}}

\renewcommand{\eqref}[1]{Eq. (\ref{#1})}

\DeclareMathOperator*{\argmax}{argmax}

\begin{document}

\twocolumn[
\icmltitle{Transformers for Supervised Online Continual Learning}

\icmlsetsymbol{equal}{*}

\begin{icmlauthorlist}
\icmlauthor{Jorg Bornschein}{dm}
\icmlauthor{Yazhe Li}{dm}
\icmlauthor{Amal Rannen-Triki}{dm}
\end{icmlauthorlist}

\icmlaffiliation{dm}{Google Deepmind}

\icmlcorrespondingauthor{Jorg Bornschein}{bornschein@google.com}

\icmlkeywords{Machine Learning, ICML}

\vskip 0.3in
]

\printAffiliationsAndNotice{}  %

\begin{abstract}
Transformers have become the dominant architecture for sequence modeling tasks such as natural language processing or 
audio processing, and they are now even considered for tasks that are not naturally sequential such as image classification. 
Their ability to attend to and to process a set of tokens as context enables them to develop in-context few-shot learning abilities. 
However, their potential for online continual learning remains relatively unexplored.
In online continual learning, a model must adapt to a non-stationary stream of data, minimizing the cumulative next-step prediction loss. We focus on the supervised online continual learning setting, where we learn a predictor 
$x_t \rightarrow y_t$ for a sequence of examples $(x_t, y_t)$.
Inspired by the in-context learning capabilities of transformers and their connection to meta-learning, we propose a method that leverages these strengths for online continual learning. 
Our approach explicitly conditions a transformer on recent observations, while at the same time 
online training it with stochastic gradient descent, following the procedure introduced with Transformer-XL. 
We incorporate replay to maintain the benefits of multi-epoch training while adhering to the sequential protocol.  
We hypothesize that this combination enables fast adaptation through in-context learning and sustained long-term improvement via parametric learning.
Our method demonstrates significant improvements over previous state-of-the-art results on CLOC, a challenging large-scale real-world benchmark for image geo-localization.
\end{abstract}

\section{Introduction}

We consider the problem of online supervised learning with modern deep neural networks, 
sometimes also referred to as online continual learning. 
Given a sequence of observations $x_t$ (e.g. images) and corresponding prediction targets $y_t$ 
(e.g. labels), we aim to find a predictive model $p(y_t | x_t, \calD_{<t})$ that minimizes the cumulative log-loss
$\calL = \sum_{t=1}^{T} -\log p(y_t | x_t, \calD_{<t})$ 
\footnote{
Other loss-functions are feasible, for example the cumulative zero-one loss which would amount to minimizing 
the total number of prediction mistakes on the sequence. 
}.
We use $\calD_{<t} = \{ (x_1, y_1), \cdots (x_{t-1}, y_{t-1}) \}$ to denote the sequence of observations and 
prediction targets up to (including) $t{-}1$.
This formulation is quite general and allows for a wide variety of approaches to solve the problem. It is also theoretically well 
motivated and encourages us to find models with many desirable properties such as fast adaptation, sample-efficient learning, 
and maintaining plasticity.
Furthermore, minimizing the cumulative next-step log-loss is a form of the Minimum Description Length (MDL) principle. 
Appendix \ref{sec:mdl} provides a deeper explanation of this connection and highlights some of its properties and benefits.

Traditionally, in online learning, the conditioning on previous examples comes from the learned model parameters $\theta$
and the model is otherwise independent across examples. Thus, the general objective is reformulated as 
$\calL = \sum_{t=1}^{T} -\log p(y_t | x_t, \theta_{t-1})$, where the model parameters are a function of the previously
observed data $\theta_{t-1} = \hat{\theta}(\calD_{<t})$. 
Often, constraints are placed on how the parameter estimator can access previously observed data.
In the streaming online-learning scenario, for instance, only the most recent example is used to update the 
model parameters: $\theta_t = \text{update-step}(\theta_{t-1}, x_t, y_t)$. For online gradient descent this simplifies to
$\theta_t = \theta_{t-1} + \alpha_t \, \nabla_\theta \log p(y_t | x_t, \theta_{t-1})$.
Within the deep-learning community, it is common to use experience replay or 
core-sets to perform multiple 
gradient steps on buffered examples to update the parameters
\citep{chaudhry2019continual}. %
Recent work argues that storing all previous data $\calD_{<t}$ is often not a major technical challenge 
and that instead of limiting the size of the replay-buffer, limits on the compute requirements 
should be considered \citep{prabhu2023online}. %
To that end \citet{bornschein2022sequential} propose to replay the sequence in-order from long-term storage.

Another line of work considers memory- or retrieval inspired approaches. These methods 
often rely on pre-trained feature extractors and similiarity measures to retrieve relevant
information from previously observed examples. 
\citet{prabhu2023online} for example apply approximate online k-nearest neighbor (kNN) to 
CLOC, a large-scale and real-world online geo-localization sequence. 
For each new observation $x_t$ they retrieve a similar example 
$(x_i, y_i)$ with $i{<}t$, and make a label prediction based on $y_i$.  

In this work we propose a hybrid approach: 
We use a transformer model $p(y_{t+1} | x_{t+1}, \calD_{(t-C) \cdots t}, \theta)$ 
that is explicitly conditioned on the $C$ most recent observations 
while at the same time training the model with online gradient 
descent on the chronologically ordered sequence $\{ (x_1, y_1), \cdots (x_{T}, y_{T}) \}$.
We use replay-streams \citep{bornschein2022sequential} 
as a simple and effective way to implement chronological replay of previously observed examples 
-- essentially re-introducing the benefits of multi-epoch training while
strictly adhering to the online next-step evaluation protocol.

The motivation behind this approch is that gradient based learning has proven to 
be effective for stationary and slowly changing data. It promises steady 
learning progress even after seeing millions of examples.
Explicitly conditioned transformer models on the other hand 
have demonstrated excellent performance when learning from short-term 
correlations within the window of the $C$ most recent observations.
Trained on data that exhibits meta-learning characteristics, 
transformers develop impressive in-context few-shot learning abilities.
We conjecture that our approach can combine the complimentary 
strengths of these two approaches: 
In-context learning to enable rapid adaptation to re-occurring but potentially sudden changes in the data sequence; and
parametric learning with SGD to enable sustained progress  over long sequence lengths,  far beyond the $C$ most recent observations.

\paragraph{Contributions.}
\begin{itemize}
\item We empirically investigate the behaviour and properties of online trained 
transformer models when applied to non-stationary supervised image classification problems.
\item We propose the priviledged information (pi) transformer, 
a variant of the standard transformer architecture that is 
tailored to sequential supervised prediction problems. We show that this architecture
often achieves faster, more stable, and better prediction performance than the standard architecture
while at the same time being more compute efficient.
\end{itemize}
Our primary evaluation is on CLOC, a large scale and real-wold online continual learning benchmark. 
We show that both the standard transformer architecture, and especially the pi-tansformer, 
achieve excellent prediction performance and far exceed previously reported prediction accuracies.

\section{Architecture and Method}

We experiment with two different model architectures: 

\begin{enumerate}[label=\arabic*),leftmargin=*]

\item 2-token approach: We use a plain decoder-only, causal transformer 
and present each example $(x_t, y_t)$ as two consecutive tokens. The underlying 
transformer processes a sequence of length $2T$. We ignore the prediction 
loss on the $x_t$ tokens: the model is trained to predict the $y_t$ tokens only.

\item Privileged information (pi) transformer: For each example 
we present $x_t$ as input token. But modify the basic transformer 
block such that each token $x_t$ also receives additional privileged information $y_t$. 
The architecture is setup so that the prediction 
at time t can not access the information in $y_t$, however can access $y_{<t}$.
Specifically, projections of $y_i$ are added to the attention keys and 
values only. We also ensure that the attention mask 
has a zero diagonal, thus prohibiting the attention heads at time $t$ to access 
values from time $t$. See \ref{app:pit} for the exact formulation.

\end{enumerate}

For both architectures the input $x_t$ is in our case the feature vector corresponding to an input image. 
We use ConvNets, ResNets and Vision Transformers (ViT) as feature extractors. 
We pick these because they are popular architectures in related works;   
we did not perform a systematic search for best performing feature extractors.

We perform Transformer-XL style \citep{dai2019transformer} online training: 
The forward- and backward pass operates on a relatively small number of 
sequential tokens, typically $S \approx 100$. 
$S$ is the chunk-length and with online SGD (without replay) we perform $T/S$ 
gradient steps when fitting a sequence of length $T$.
The self-attention heads, however, operate on a larger buffer of attendable keys and values. 
We store the keys and values for the $C$ most recent tokens in a KV-cache ringbuffer.
$C$ is the the (sliding) attention window size of the architecture.
We use Multi Query Attention (MQA, \citet{shazeer2019fast}), and thus only store
one key- and one value vector per transformer block and token. 

With a 8-block deep transformer architecture and $dim_k = 128$ 
we store $8 {\cdot} 128 {\cdot} 1024 \approx$ 1Mi floats to 
keep $C{=}1024$ recent tokens accessible.
We use MQA and MLP layers in a parallel configuration, following
some recent models like CodeGen, PaLM, and GPT-NeoX.

\tikzstyle{op}=[rectangle,minimum height=0.65cm,draw]
\tikzstyle{value}=[rectangle]

\subsection{Replay-streams: In-order Replay}

We use the approach from \citet{bornschein2022sequential}
and introduce a hyper-parameter {\em num-streams} ($E$) that controls the replay factor, 
and therefore the number of effective epochs:
We maintain $E$ separate sequence states: for each its own KV-cache and
its own deterministic, sequential data-reader that yields the examples
in deterministic order. 
With each turn, all streams are advanced and 
perform a gradient step on chunk size $S$ many examples.
The first stream will proceed steadily through the data-sequence from 1 to $T$ and suffer the (cumulative) log-loss and prediction error which we report. 
The remaining $E{-}1$ streams replay data that has been previously seen by the first one.
By stochastically resetting some of the $E{-}1$ replay data-readers and their associated KV-caches 
back to position 0 we achieve, in expectation, uniform replay of the past without requiring large in-memory replay-buffers.  
See algorithm \ref{alg:replay} for details.

Due to the approximate uniform replay, our method resembles a particular form of meta- or bi-level learning:
At time $t$, the {\em outer-learner} optimizes the parameters $\prm_t$ to improve the prediction $p(y_{i+1} | x_{i+1}, \calD_{i-C \cdots i}, \prm_t)$ for all $i \le t$ simultaneously; 
while the inner, fast learner~ \citep{schlag2021linear} is non-parametric and relies purely on in-context learning with $C$ tokens in its attention window.

\begin{lstfloat}[t]
\begin{lstlisting}[language=Python]  
cumulative_nll = 0.
data_readers = [
    new_data_reader() for _ in range(num_streams)]
kv_caches = [
    empty_kv_cache() for _ in range(num_streams)]

# Iterate over all data
for pos in range(num_examples):
  # Read new data, perform gradient step 
  # and accumulate NLL.
  data = next(data_readers[0])
  nll, kv_caches[0], model_params = gradient_step(
    data, kv_caches[0], model_params)
  cumulative_nll += nll
  
  # Read data from remaining streams for replay.
  for s in range(1, num_streams):
    data = next(data_readers[s])
    _, kv_caches[s], model_params = gradient_step(
            data, kv_caches[s], model_params)
    
    # Reset each replay stream with probability 1/pos
    if bernoulli(1/pos):
        data_readers[s] = new_data_reader()
        kv_caches[s] = empty_kv_cache()
\end{lstlisting}
\vspace{-0.25cm}
\caption{
    Replay-streams training:
    \texttt{\small new\_data\_reader()} creates a deterministic data-reader 
    that yields successive examples. %
    \texttt{\small gradient\_step($\cdots$)} performs a training step and returns 
    the log-loss, state of the sequence model (KV-cache) and the updated parameters. 
    In practice we perform the algorithm with chunks of $S$ successive examples 
    instead of individual ones. The reset probability to approximate uniform replay 
    for chunk-size $S$ is $\frac{S}{t}$ instead of $\frac{1}{t}$. 
    \label{alg:replay}
}
\end{lstfloat}

\section{In-context and Meta-learning} %

Recent work has pointed out that the in-context learning abilities of transformer models can be 
understood and enhanced when taking a meta-learning perspective: 

Different works have reported several relevant results while studying online learning and non-i.i.d. data and their relation to meta-learning and in-context learning and transformers \citep{Lee2023-hn}. \citet{ortega2019meta} shows that meta-learning of memory-based models leads to near Bayes-optimal solutions. The authors also highlight that meta-learning can occur spontaneously through online learning when the model capacity is bounded and the data is produced by a single generation process. They however warn against the challenges that this type of emergent meta-learning can lead to, due to the lack of control over it. Safety problems can arise, and need to be taken into consideration. \citet{mikulik2020meta} and \citet{genewein2023memory} reinforce the previous results both theoretically and empirically: The former shows that not only meta-learned models with memory converge to Bayes optimal agents, they are also indistinguishable from the perspective of predictions and computations. The latter shows that the meta-learning of memory-based models on non-stationary distributions leads to Bayes-optimal predictors. \citet{chan2022data} and \citet{Singh2023-qf} take a difference stance and study the emergence of in-context learning behavior in transformers through the lens of data distribution. \citet{chan2022data} highlight the importance of both the architecture (i.e. the use of attention mechanisms) and the data distribution for this behavior to appear. The authors show that the few shot learning capabilities of transformers arise when data is `bursty' and appear in clusters rather than uniformly across the training data, when data is highly skewed and the classes follow a Zipfian distribution, and when the meaning of labels is dynamic and context dependent. We note that all these characteristics make the transformer training divert from  the classical i.i.d. setting, and gets closer to a non-stationary online setting, which connects to the results in \citet{ortega2019meta}. These papers also distinguishes between in-context learning and in-weights learning. \citet{Singh2023-qf} highlight that the in-context learning behavior is transient, and that in-weight learning reemerges if the model is overtrained. 

We note that none of these papers studies transformers in the online learning scenario, 
but they describe a perspective on in-context learning with transformers that inspired our approach here.
In particular, we bridge the gap between in-context and in-weight learning, highlighting cases where they can work in synergy. 
We think that such synergy could be instrumental in solving the prospective learning problem discussed in~\citet{de2023prospective}.

\begin{figure*}[t]
    \centering
    \includegraphics[width=0.45\linewidth]{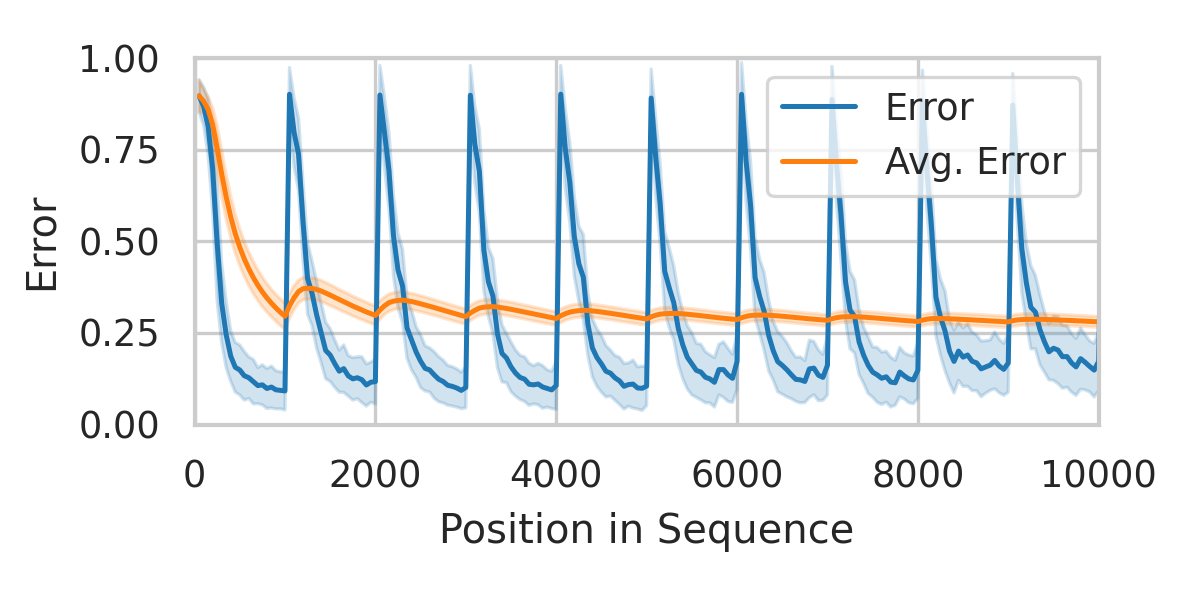}
    \hspace{-0.4cm}
    \includegraphics[width=0.45\linewidth]{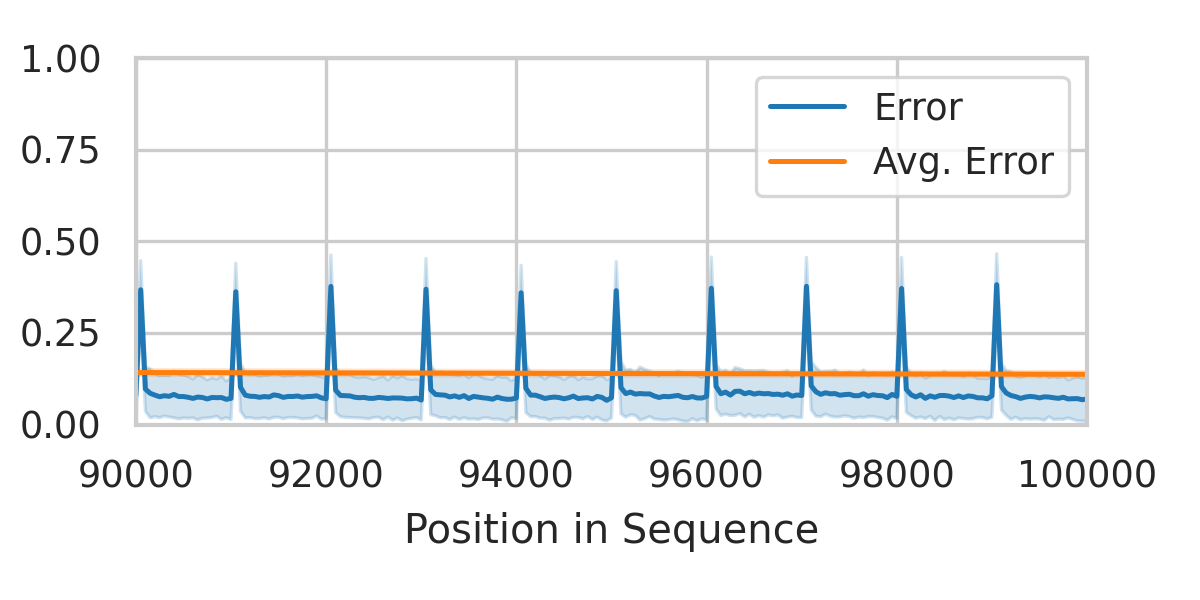}
    \vspace{-0.4cm}
    \caption{
      Instantaneous and averaged prediction performance for the first and last 10 tasks of Split-EMNIST: 
      Image-to-label mappings are constant within each task, but randomly reassigned at task boundaries    
      (averaged over 200 data generating random seeds).
    \label{fig:emnist-plots}
    }
\end{figure*}

\begin{figure*}[]
    \centering
    \includegraphics[width=0.45\linewidth]{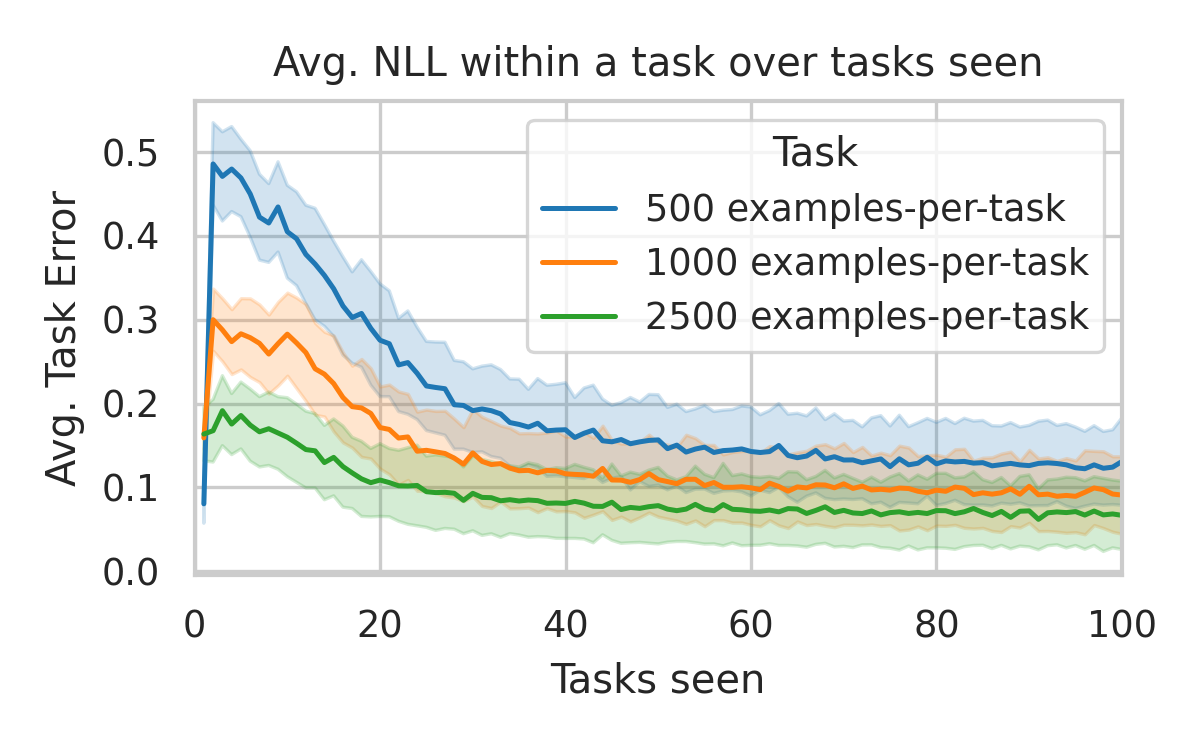}
    \includegraphics[width=0.45\linewidth]{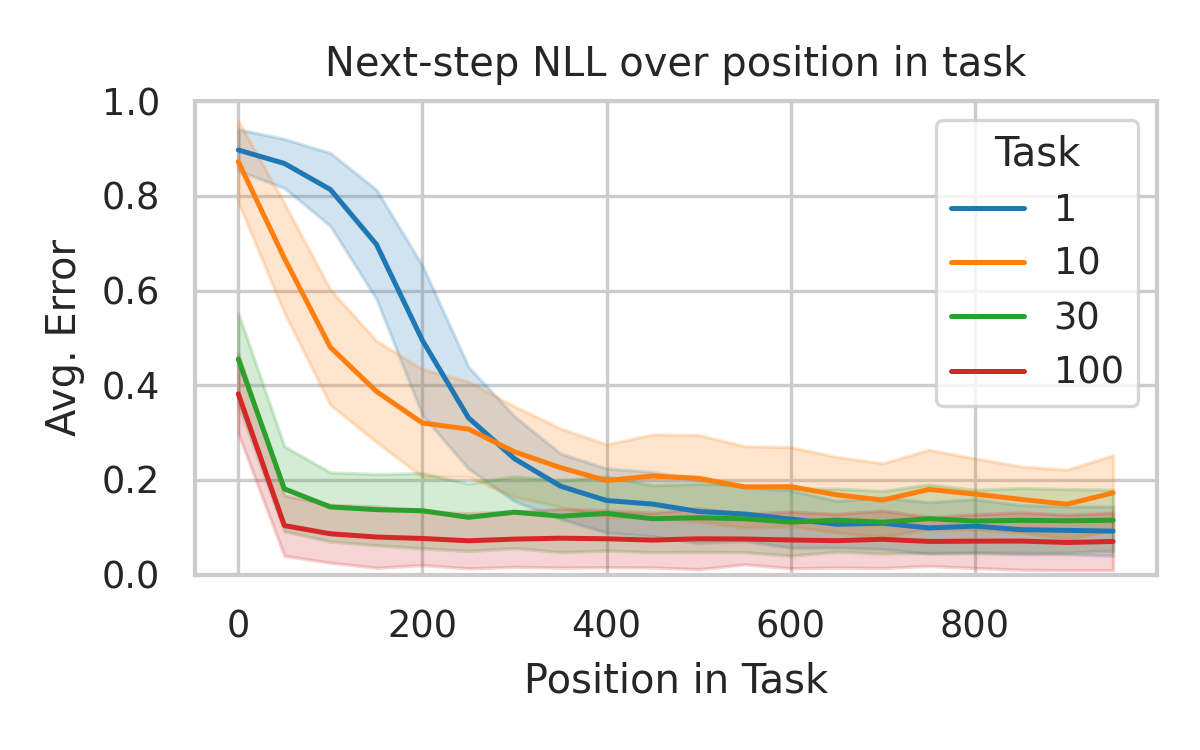}
    \vspace{-0.3cm}
    \caption{
    Alternative visualization of the Split-EMNIST experiments from Fig. \ref{fig:emnist-plots}:
    {\bf Left:} Average performance per task shows strong forward-transfer after struggling during the first 10 to 20 tasks.
    {\bf Right:} Detailed look at the within-task performance for the scenario with 1000 examples per task. 
    The model is a strong few-shot learner after seeing 30 tasks and further improves until at least task 100 
    \label{fig:emnist-detail}
    }
\end{figure*}

\begin{figure}[h]
\includegraphics[width=0.9\linewidth]{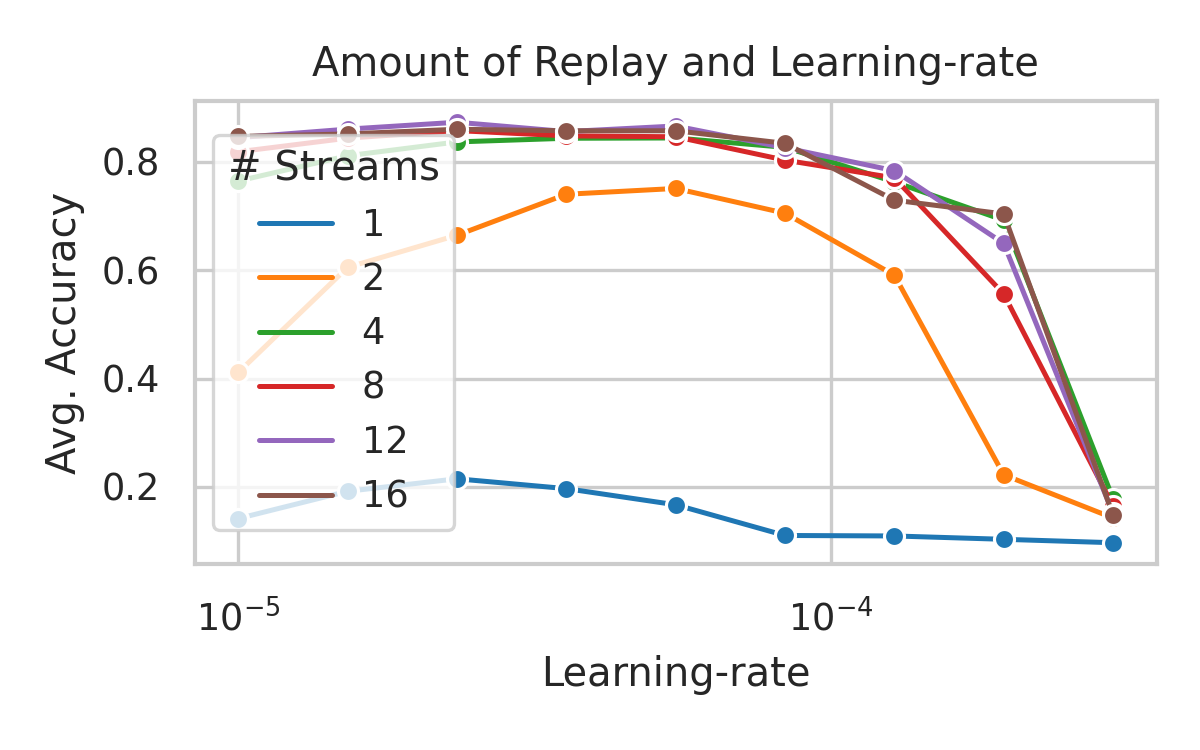} \\
\vspace{-0.5cm}
\caption{Average accuracy at the end of the sequence for different amount of replay (epochs) 
and learning-rates on Split-EMNIST (1000 examples per task, 100 tasks).
\label{fig:emnist-num-streams}
}
\end{figure}

\begin{figure*}[t]
    \centering
    \includegraphics[width=1.0\textwidth]{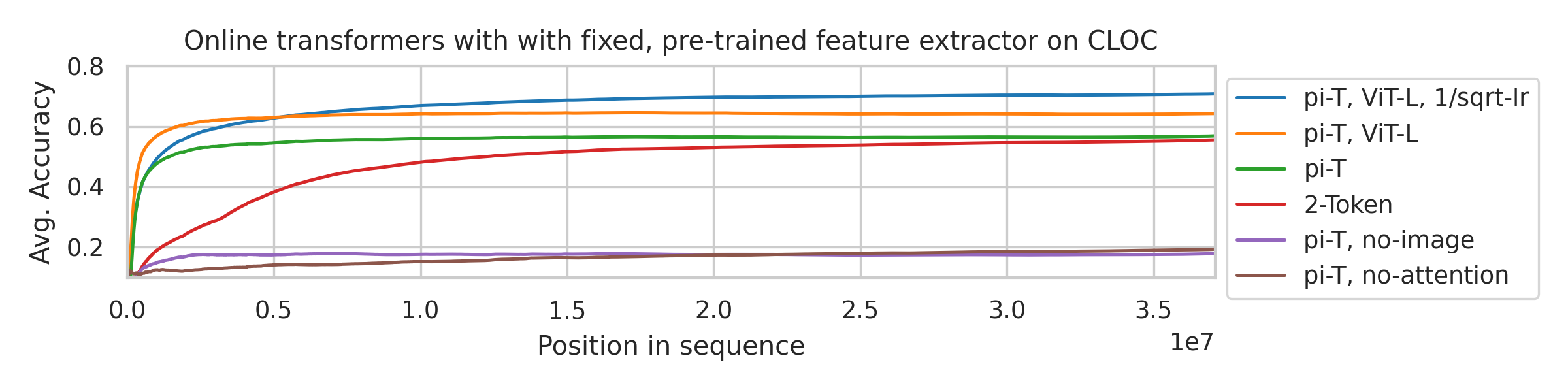} \\
    \vspace{-0.7cm}
    \caption{CLOC with pretrained and frozen feature extractors. 
    We show the best performing models (in terms of final avg. accuracy) 
    from the hyper-parameter cube in \ref{app:cloc-features-hypercube}
    We also show pi-Transformer ablations:
    either without input features $x_t$, or without attention ($C=0$).
    }
    \label{fig:cloc-features-seq}
\end{figure*}

\section{Experiments on Synthetic Toy Data}\label{sec:emnist_exp}

\paragraph{Piece-wise stationary Split-EMNIST.}
We create synthetic, piece-wise stationary sequences from underlying labeled data sets.
The sequences are composed of 100 tasks. Each task is a 10-way classification problem 
and is generated by first randomly selecting 10 classes from the underlying dataset, 
assigning them to the 10 observed label-classes and randomly selecting corresponding images
until the desired number of examples for this task have been collected. From the perspective 
of the (task-agnostic) online-learner, a task boundary is a sudden event when the distributions of 
the observed data $x$ and the associated target labels $y$ change abruptly. 
Even if similar inputs $x$ have been observed before, they are now with high probability assigned to different labels $y$.
We create Split-EMNIST from the balanced EMNIST data set \citep{cohen_afshar_tapson_schaik_2017}, 
which provides 47 underlying classes.

We use the 6-layer convolutional neural network described by \citet{blier2018description} as a feature extractor, 
however double the number of channels in all stages. 
On top of that we use 4 transformer blocks with a backbone width of 256 units.
We first study the behaviour of the pi-transfomer architecture.
We choose a chunk size $S$ of 50, a constant learning rate of 1e-4 and 8 replay-streams. 
See Appendix \ref{app:emnist} for details of the model architecture and hyperparameters. 
Figure \ref{fig:emnist-plots} shows the next-example prediction performance averaged over 200 
random seeds for the data generation. Averaging over many data generating seeds ensures
we plot clear trends even though there is high across example- and across task variability. 
Note however that the {\em average} accuracy has very low variability, even for a single run. 
Figure \ref{fig:emnist-detail} shows a detailed view of the average NLL per task, or per 
position within a task. We observe that after initially struggling during the first $\approx$ 20 tasks,
the model learns to be a highly efficient few-shot learner that only requires a few examples at the 
beginning of each task to make accurate predictions.
Figure \ref{fig:emnist-num-streams} shows the effect of the number of replay streams (epochs) and learning rate on the final result. 
We observe that without replay, the model usually does not learn to use in-context information effectively to make predictions.
With replay, the model is forced to learn one set of parameters $\prm_t$ that work at $t+1$,
and at the replay positions $< t$. It is thus strongly encouraged to be an in-context learner. 

We repeat the experiments with the 2-token architecture and observe generally the same behaviour. 
However, the 2-token approach is learning slower in the sense that it requires more samples and more tasks to achieve 
the same performance. 

In Appendix \ref{app:cifar} we presents analogous experiments with sequences based on 
CIFAR-100. We generally observe the same behaviour, however we use a deeper feature extractor
and it requires more replay and more examples to obtain good few shot performance and is more sensitive to hyper-parameter choices. 
We note that these results confirm the observations previously highlighted in~\citet{lesort2022challenging}, but with a different replay strategy.

\section{Large scale continuous geo-localization}
\label{sec:cloc}  %

Our primary evaluation with real-world data is on CLOC, the continual geo-localization sequence introduced 
by \citet{cai2021online}. 
It consists of $\approx 39M$ chronologically ordered images with their geo-location as a categorical prediction target. 
Figure 2 in \citep{cai2021online} shows that the data is strongly non-stationary, 
and that traditional i.i.d. training results in a held-out accuracy between 10 and 20\%.
We failed to download or decode around 5\% of the images, 
which leaves us with a sequence of 37,093,769 images. 
Different protocols have been used to train and evaluate models. Most notable differences include 
a) whether pre-trained feature extractors are used; 
b) whether the feature extractor is fine-tuned.

We obtain substantial improvements over previous state-of-the-art for the two scenarios we consider here: Learning with frozen pre-trained feature extractors and learning the whole model from scratch. 
In both cases we almost double
the archived average accuracy (Table \ref{tab:cloc}). 
Note that initial results from \citet{cai2021online} considered a filtered versions of the task
where some short term correlations are explicitly removed; 
recent works \citep{titsias2023kalman, bornschein2022sequential,prabhu2023online} however focus on the unfiltered average online next-step accuracy. 
In the following we describe the experiments for the frozen pre-trained feature-extractor and online-learned feature extractor from scratch separately.

\subsection{Pre-extracted image features.} 
\label{sec:cloc-features}

\begin{table}[h]
    \centering
    \caption{Results on CLOC: 
        Experience Replay (ER, \citet{cai2021online}), 
        ACM \citep{prabhu2023online}, 
        Replay Streams \citep{bornschein2022sequential}, 
        Kalman Filter \citep{titsias2023kalman}.
    }   
    \label{tab:cloc}
    \small
    \begin{tabular}{lcccc}
    \toprule
    Method & Pretrained & Finetuning & Avg. Acc. \\
    \midrule
    ER & \checkmark & \checkmark & 20\% \\
    ACM & \checkmark & -  & 26\% \\
    Replay Streams & \checkmark & - & $\approx$ 38\% \\
    Replay Streams & - & \checkmark & $\approx$ 37\% \\
    Kalman Filter & \checkmark & - & 30\% \\
    Kalman Filter & - & \checkmark & 37\% \\
    \hline
    Ours (no-image) & - & -  & 18\% \\
    Ours (no-attention) & \checkmark & -  & 19\% \\
    \hline
    {\bf Ours} (Superv. ResNet) & \checkmark & -  & {\bf 59\%} \\
    {\bf Ours} (MAE ViT-L) & \checkmark & -  & {\bf 70\%} \\
    {\bf Ours} & - & \checkmark & {\bf 67\%} \\
    \bottomrule
    \end{tabular}
\end{table}

\begin{figure*}[t]
    \centering
    \includegraphics[width=0.9\linewidth]{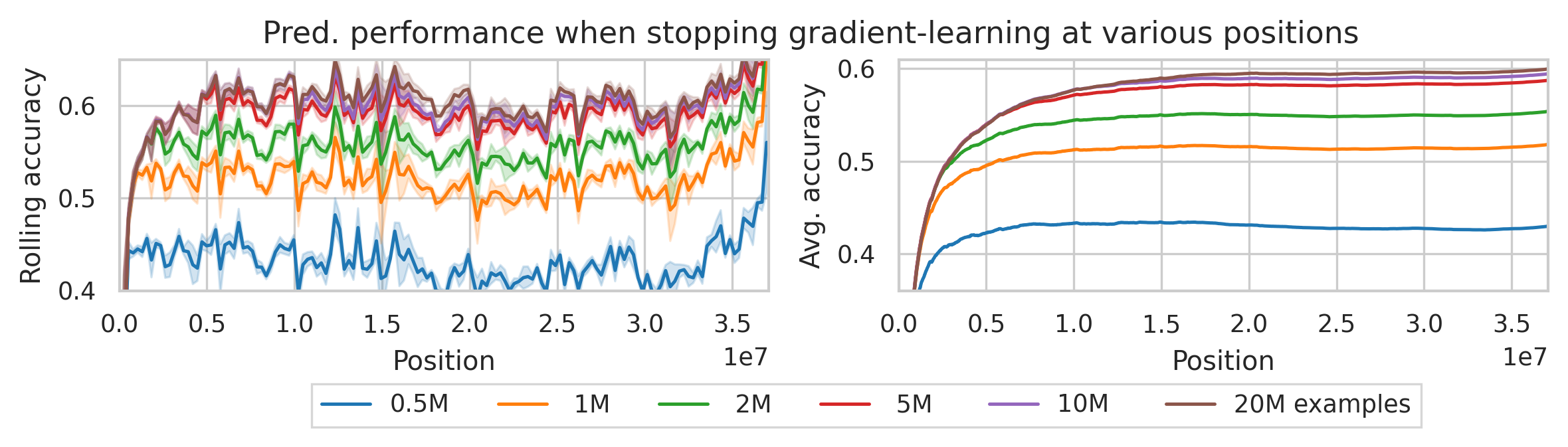}
    \vspace{-0.3cm}
    \caption{
    Stopping gradient updates at various positions to investigate the performance of {\em in-context} conditioning
    alone. Here for the pi-Transformer on CLOC with a fixed, pre-trained ResNet-50 feature extractor. 
    Gradient based online-learning is most important at the beginning of the sequence, however keeps on 
    contributing even after 10M examples.
    } 
    \label{fig:cloc-features-stopping}
\end{figure*}

\begin{figure*}[t]
    \centering
    \includegraphics[width=0.45\linewidth]{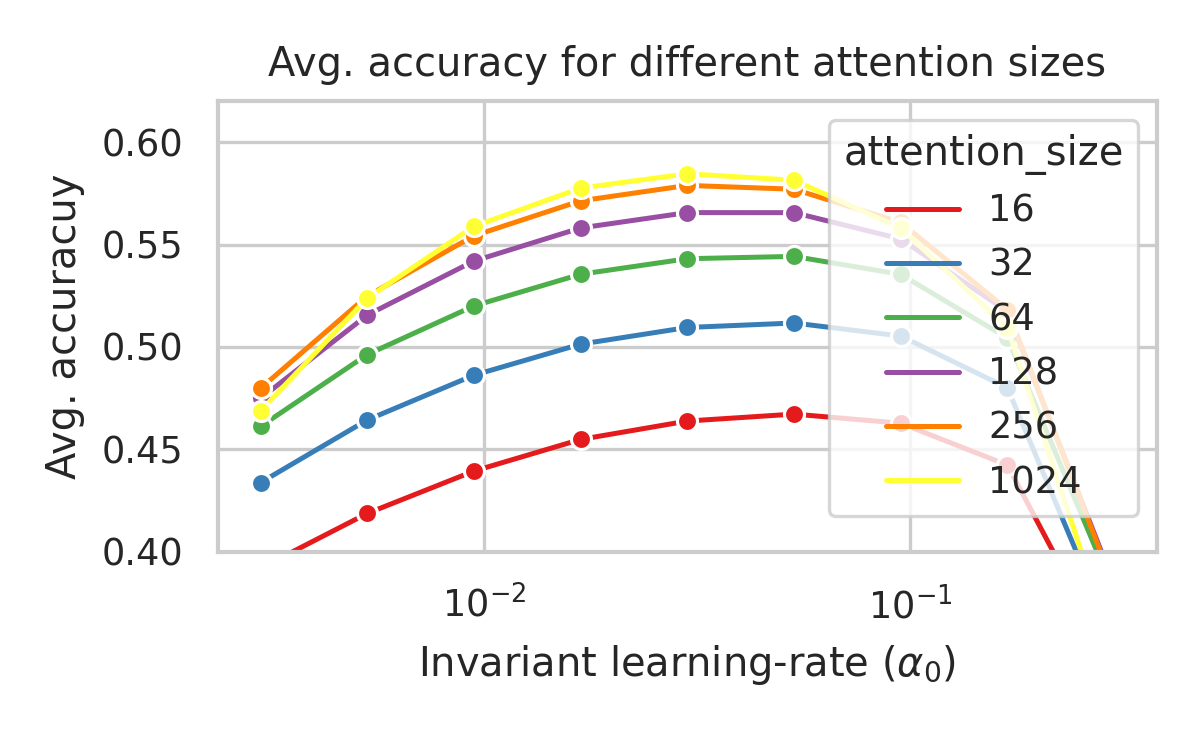}
    \includegraphics[width=0.45\linewidth]{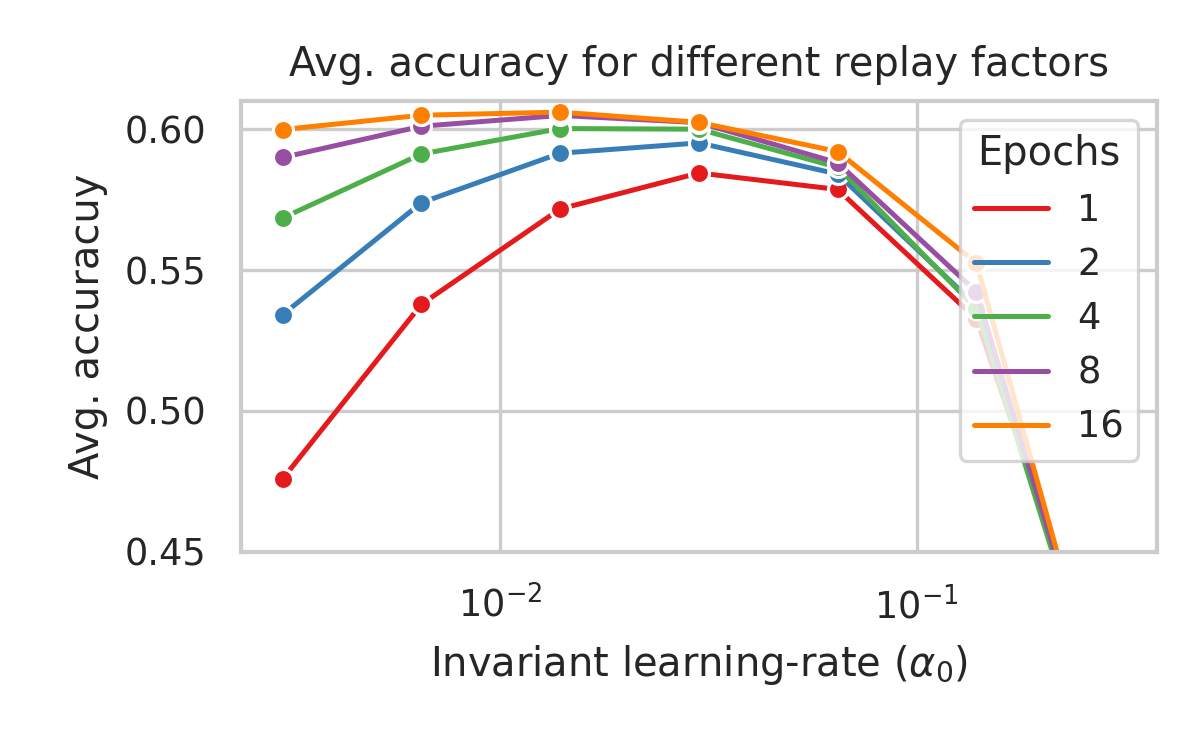}
    \vspace{-0.5cm}
    \caption{
    Pi-Transformer on CLOC with a fixed, pre-trained ResNet-50 feature extractor: 
    {\bf Left)} Increasing the attention window up to about 512 examples improves the results.
    {\bf Right)} Performing more replay gradient steps improves performance and makes the model more robust to lower learning-rates.
    } 
    \label{fig:cloc-sensitivity1}
\end{figure*}

The choice of the pre-trained and frozen feature extractor has a significant impact on the 
achievable performance. We first present the results when using the 
top-level activations of a supervised ImageNet trained ResNet-50,
just like \citet{prabhu2023online}. 
Consistent with their observation, we also obtain substantially better results when using 
self-supervised trained feature extractors such as a MAE ViT-L \citep{he2022masked}.
With pre-extracted features and depending on the amount of replay and the size of 
the transformer it requires between $10^{14}$ and $10^{17}$ FLOPs to run CLOC. 
This corresponds to walltimes between 15 minutes and 12h hours on a single GPU. 
The base-architecture for all experiments is a 4-block deep transformer 
with a backbone width of 1024 units. 
We set the attention window size to 512 and choose a chunk size $S$ of 256 examples 
with a constant learning rate for the AdamW optimizer.
Figure \ref{fig:cloc-features-seq} shows typical avg. accuracy curves.
We include additional baselines: 
a) an online transformer that has no access to the image $x_t$  
and that makes prediction from the label sequence alone;
b) an online transformer without attention, i.e., context size $C{=}0$
that relies on the current image features alone; 
and c) an oracle predictor, that predicts the correct location whenever 
an image with the same location was observed during the previous 100 examples.

We run extensive hyper-parameter sweeps to characterize the model's performance.
For these we first focus on the pi-transformer architecture. 
Results for the 2-token approach can be found in the Appendix \ref{app:cloc-features-2t}.
We observe that the optimal (constant) learning rate is typically $\alpha \approx \frac{\alpha_0}{D}$, 
where $D$ is the width of the transformer architecture and $\alpha_0~\approx~ 3{\cdot}10^{-2}$. 
This is consistent with the results from \citet{yang2021tuning}.
Figure \ref{fig:cloc-sensitivity1} (left) shows the average accuracy at the end of the sequence for various 
attention window sizes as a function of the learning rate.
The plot on the right shows the influence of the number of replay streams (epochs).
While more replay generally improves the predictive performance, we obtain already decent results
completely without replay. 
Figure \ref{fig:cloc-sensitivity2} in the Appendix shows the influence of the model size (width)   
and of the chunk size $S$ on the final performance. The choice of the chunk-size is not critical to 
obtain good results, however it does correlate with the optimal learning rate, just like batch-sizes 
do for regular iid. training. 
Overall, we observe that learning a pi-transfomer on top of pre-extracted features is very robust and 
well behaved under hyper-parameter variations. 

Our approach combines in-weight and in-context learning into an online-learning algorithm where these two mechanism 
are synergistically intertwined. 
To get a sense of their relative contributions we run two ablations: First, we train a model with $C{=}0$, 
effectively disabling the attention mechanism in the transformer. We obtain an average accuracy of only about $19 \%$ 
-- comparable to previously reported results for experience replay (ER, see Figure \ref{fig:cloc-features-seq}). 
Secondly, in Figure \ref{fig:cloc-features-stopping}, we train online transformers and disable gradient updates once the learner reaches 
positions 0.5M, 1M, 2M, 5M, 10M or 20M. 
From these points onward the weights are frozen and the predictor relies on in-context conditioning alone. 
As one would expect, parametric learning is most important at the beginning of the sequence, however still contributes to 
improved results after > 10M datapoints. 

We run the same suite of experiments with the 2-token approach.  
Learning curves often show distinct, non-smooth improvement steps (Figure \ref{fig:cloc-2t-phase-changes}). 
\citet{olsson2022context} associate such step-improvements with 
phase-changes in the learning dynamic and with the formation of induction-heads that are 
crucial for in-context learning. 
The exact occurrence of these steps is stochastic and sensitive to hyper-parameter choices. 
The pi-transformer learning curve shows in general more reliable improvements, 
presumably because the superimposed label information provides an adequate inductive bias 
where the label association does not have to be learned. 
Appendix \ref{app:cloc-features-2t} presents more results for the 2-token approach.

In Figure \ref{fig:cloc-attn-sensitivity} we directly compare the sensitivity of the different 
transformer based approaches to changes of the the attention-window size $C$.
Figure \ref{fig:cloc-replay-sensitivity} shows the sensitivity to the number
of replay-streams (epochs) $E$.

\begin{figure}[h]
    \centering
    \includegraphics[width=1.0\linewidth]{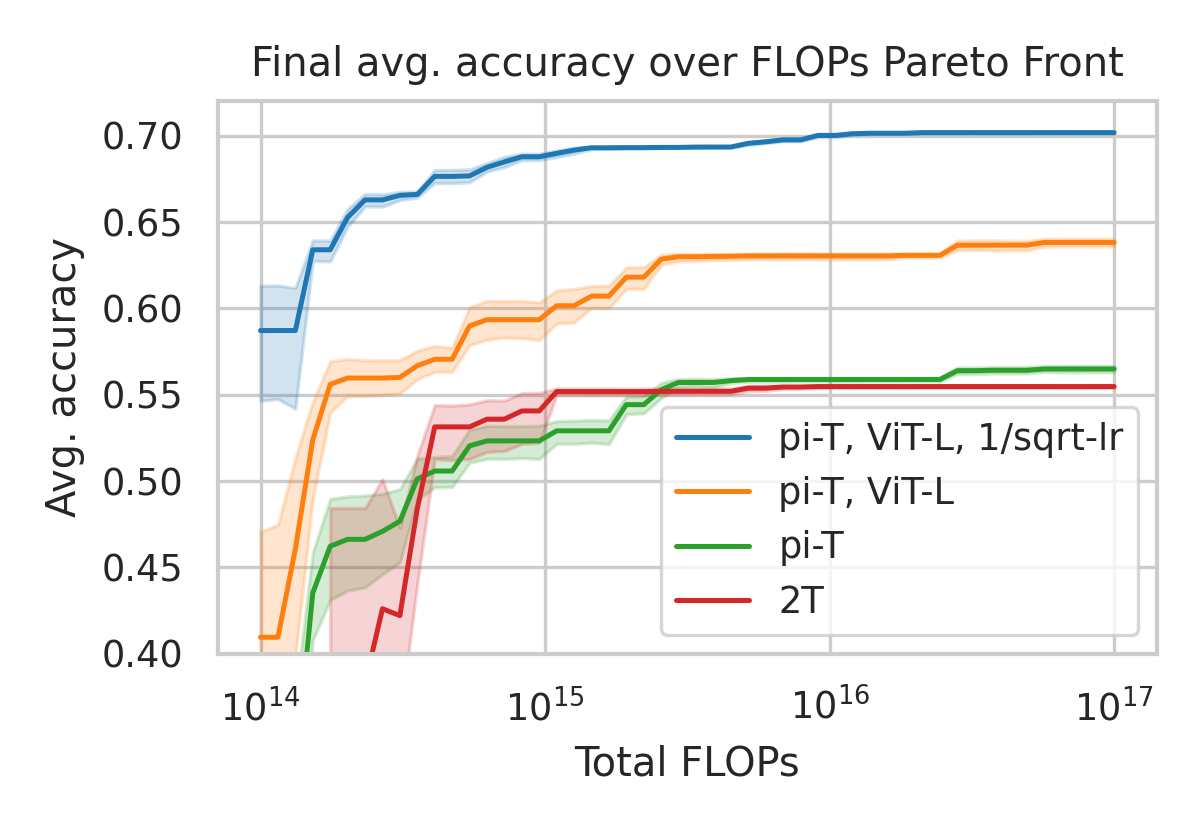} 
    \\
    \vspace{-0.5cm}
    \caption{
    Pareto fronts for online learners with fixed, pre-trained feature extractors on CLOC: 
    Final avg. accuracy at the end of the sequence over total number of FLOPs (MACs).
    Excluding the cost of the feature extractor.
    We run a broad hyper-parameter cube with 250 experiments for each method to 
    obtain the pareto fronts.
    } 
    \label{fig:cloc-pareto}
\end{figure}

\begin{figure}[h]
    \centering
    \includegraphics[width=0.95\linewidth]{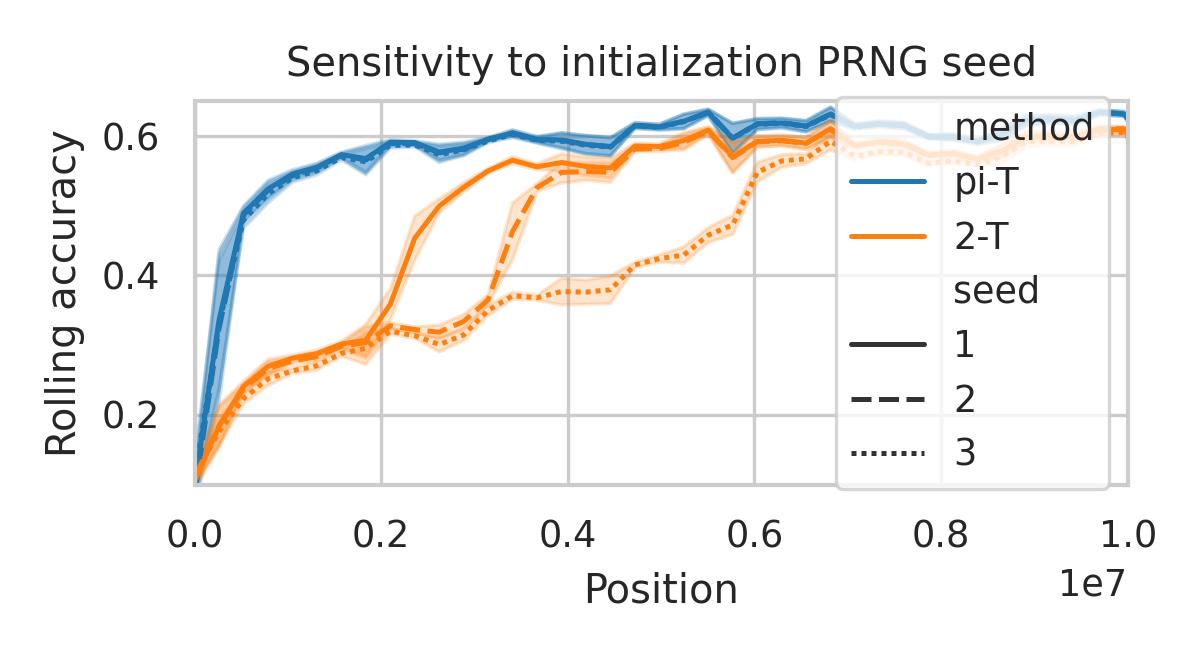}
    \vspace{-0.5cm}
    \caption{
    First 10M examples of CLOC with fixed, pre-extracted features: The pi-Transformer shows a smooth 
    accuracy improvements. Accuracies for the 2-token approach exhibit discrete 
    step improvements with their exact location depending on the random initialization seed.
    }
    \label{fig:cloc-2t-phase-changes}
\end{figure}

\begin{figure}[h]
\centering
\includegraphics[width=0.95\linewidth]{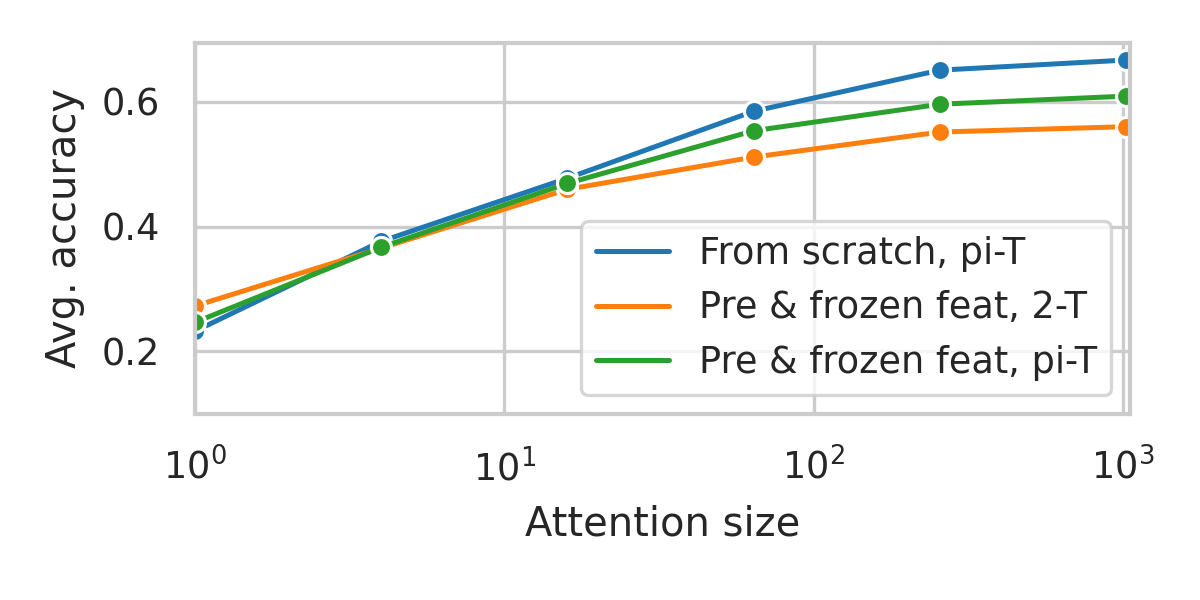}
\vspace{-0.5cm}
\caption{
Sensitivity to the size of the attention window $C$ for different models. 
For each experiment we sweep over 7 different transformer learning rates and report the best run. 
}
\label{fig:cloc-attn-sensitivity}
\end{figure}

\begin{figure}[h]
\centering
\includegraphics[width=0.95\linewidth]{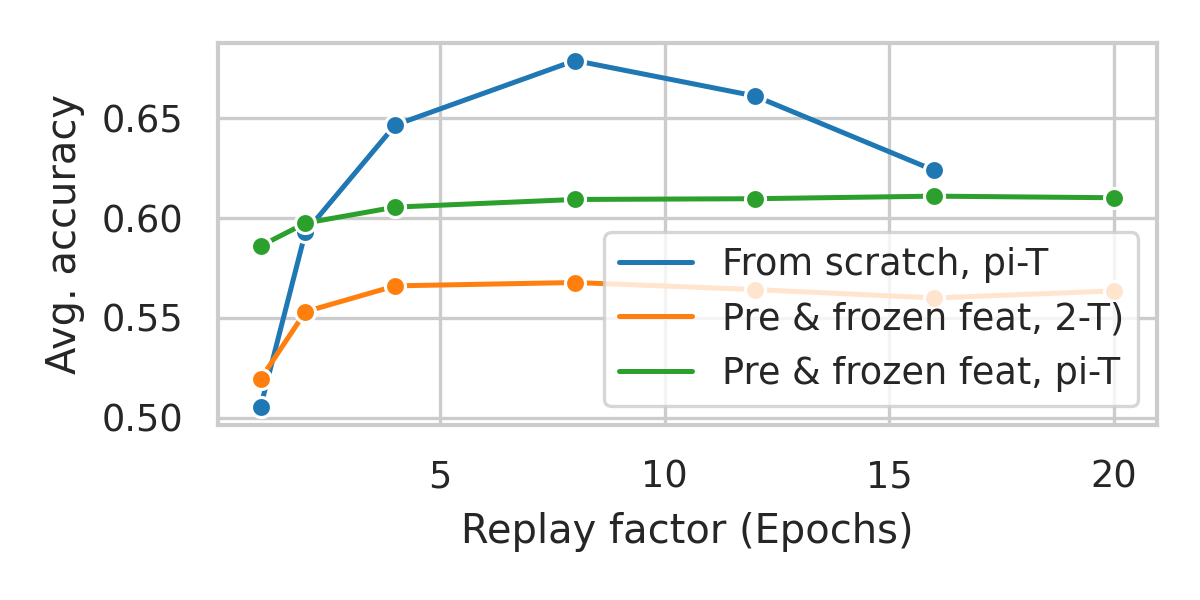}
\vspace{-0.5cm}
\caption{
Sensitivity to the number of replay-streams (epochs) $E$ for different models. 
For each experiment we sweep over 7 different transformer learning rates and report the best run. 
}
\label{fig:cloc-replay-sensitivity}
\end{figure}

\paragraph{Pareto fronts.}
The total computational cost for fitting the model to the sequence varies substantially 
depending on the model size, depth, and number of replay streams. 
To compare approaches fairly, we try to characterize their pareto-front in terms of
final avg. accuracy over required FLOPs.
We define a hyper-parameter cube that spans a computational cost between 
around $10^{14}$ to $10^{17}$ FLOPs\footnote{We only count multiply-and-add operations (MACs)}
(see Appendix \ref{app:cloc-features-hypercube} for details).  
We show the results in Figure \ref{fig:cloc-pareto}.
Overall, with such a long sequence, the final average performance of the 2-token approach 
and the pi-transformer are very similar. Much larger gains can be obtained by switching to 
a different feature extractor, or adding learning-rate decay.
We note that well performing pi-transformers were often relatively shallow (2 to 4 blocks) and wide, while 
well performing 2-token models preferred more depth (8 blocks), but became narrower.

\begin{figure}[t]
    \centering
    \includegraphics[width=0.95\linewidth]{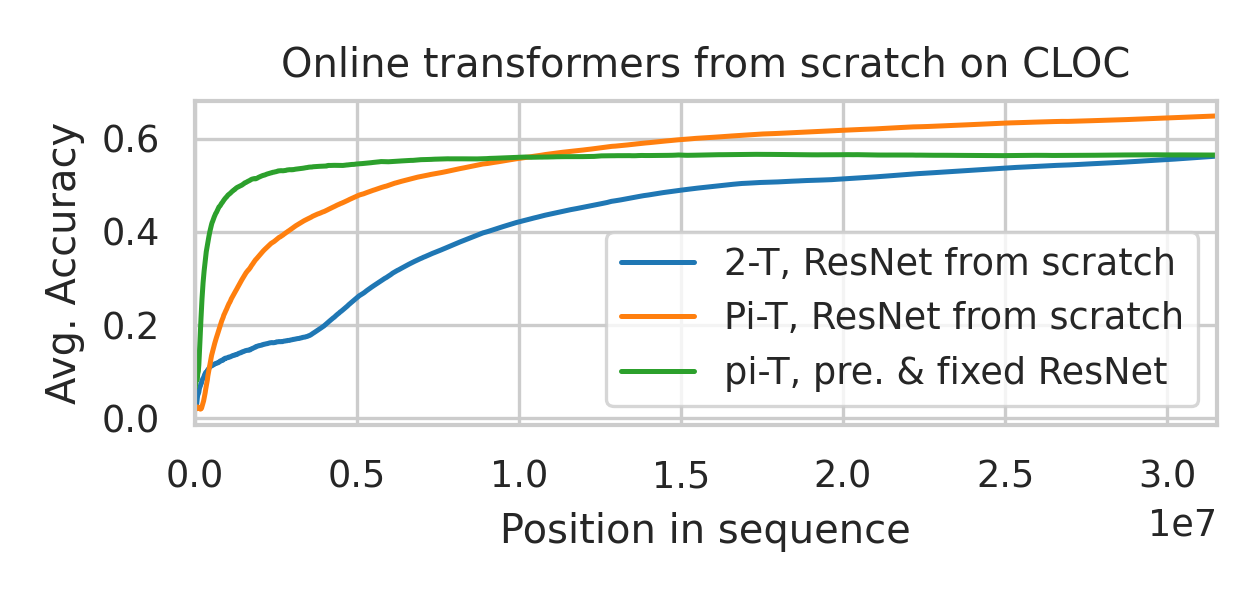} \\
    \vspace{-0.5cm}
    \caption{CLOC with ResNet feature extractor from scratch. For comparison 
    we add the learning curve for a pi-Transformer with pretrained and frozen 
    ResNet features from Figure \ref{fig:cloc-features-seq}.
    }
    \label{fig:cloc-seq}
\end{figure}

\subsection{From scratch.} 
Our second scenario for CLOC is 
to jointly online-learn the feature extractor from scratch. 
We consider ResNet style feature-extractors. However,  
off-the-shelf ResNet architectures employ batch-norm layers, which introduce 
non-causal information-leakage between successive examples due to 
the shared activation-mean within a mini-batch. 
We therefore take inspiration from \citet{brock2021high} and \citet{liu2022convnet} 
and replace the batch-norm with layer-norm instead.
The changes are consistent with the recommendations from \citet{lyle2023understanding}.
The details can be found in Appendix \ref{app:cloc-resnet}.
We observe that the the feature extractor and the temporal transformer  
prefer noticeable different learning rates in order to achieve optimal results: 
Reasonably sized ResNets prefer a learning rate between 3e-4 and 1e-3 while 
the transformer prefers again $\alpha {=} \frac{\alpha_0}{D}$ with $\alpha_0{\approx}$1e-2.
Therefore, we tune the learning rates for these two model components independently and leave 
it to future work to find a more elegant solution. We set the feature extractor learning rate 
to 3e-4 and choose a transformer backbone width of $D{=}1024$ with a 
transformer learning rate of $\alpha {=} \frac{\alpha_0}{D} {=}$ 1e-5.

Figure \ref{fig:cloc-seq} shows the average accuracy of our best performing pi-transformer and
2-token transformer architectures. We include the learning curve for a pi-transformer on 
pre-extracted ResNet-50 features for comparison. 
Figures \ref{fig:cloc-attn-sensitivity} and \ref{fig:cloc-replay-sensitivity} show the
sensitivity to varying the attention size and number of replay streams compared to models
with pre-trained and fixed feature extractors: 
Learned representations can take better advantage of large context windows $C$. 
However training is more sensitive to hyperparameters, as exemplified by the 
decline of predictive performance as a result of increasing $E$ beyond ${\approx} 10$.
We note that re-tuning the feature-extractor learning-rate allows for using more replay 
with marginal improvements for the final accuracy.

\section{Conclusions}
In this paper we study transformers in the challenging setting of supervised online continual learning: 
We combine the transformer \citep{vaswani2017attention} architecture, online-training with KV caching \citep{dai2019transformer} and
replay streams \citep{bornschein2022sequential} into a practical algorithm for online learning. 
We propose two concrete methods: 
a 2-token variant where the input features and the label are fed as two separate tokens, 
and privileged information variant where the basic transformer block is modified to receive 
additional information in the form of a projection of the labels. 

We evaluate these variants in two online learning scenarios: 
a) learning on synthetic, task-agnostic and piece-wise stationary sequences and 
b) on a naturally occuring large-scale sequence with continuous drift. 
The scenario with synthetic data allows us to study the emergent meta-learning-like behavior in detail:
As the learner progresses in the sequence, we first observe a performance drop that is generally associated 
with {\em loss of plasticity} issues. However, after encountering more task boundaries, the model learns to 
be a highly effective few-shot learner that increasingly rapidly performs well on new tasks.  
In the second scenario, on CLOC real-world data, both the 2-token and the privilledge-transformer approaches reach unprecedented levels of predictive performance; 
almost halving the state-of-the-art average error-rate with our best results.
We obtain these results for both pretrained and frozen image features, and also with feature extractors trained online from scratch.

Overall this study sheds light on interesting interactions between in-context and in-weight learning in transformers. It shows scenarios where they work in tandem to improve the performance and the efficiency of online learning methods. With this paper, we take a step towards bridging the gap between these two mechanisms, and further studies are needed to continue in this direction and understand better the synergies between them. 

\subsection{Limitations and Future Work}

There are several areas of improvements that we leave to future works: 
Training a deep feature extractor and the temporal transformer jointly 
sometimes requires two different learning rates for optimal results -- which leads to expensive and inconvenient hyperparameter searches (Section~\ref{sec:cloc}). 
We also mostly stick to constant learning rates even though initial experiments suggest that better results can be obtained 
by introducing learning rate schedules.
In  this work uses basic and potentially suboptimal pre-trained feature extractors for the images. 
We leave the study of the impact of more carefully designed embeddings to future works.

\section*{Impact Statement}

The paper presents work whose goal is to advance the field of Machine Learning. There are many potential societal consequences 
of this work, most of them are however not specific to our contribution and we think it is out of scope to discuss them here.
Beyond these universal considerations, online-learning poses some additional opportunities and challenges. 
Most notably: The predictive performance and characteristics of online-learning systems changes over time. 
This is in contrast to static models, where only the data-distribution might change and 
as a result undesirable predictions might be produced. 
With online learning, it is the model itself that changes over time too. 
If the same online-model is shared among mutually non-trusting clients, every single one of them, 
maliciously or unintentionally, might cause severe and harmful mispredictions for the others. 
We believe that no such online system should be deployed without extensive additional safety assessments.
Unfortunately, this precludes harnessing some of the benefits of online adapting systems:
continuous tracking and steady improvements in face of a changing world, and improved efficiency of the learning systems.

\bibliographystyle{icml2024}
\bibliography{references}

\appendix

\newpage
\onecolumn

\section*{Appendix}

\section{Learning on an Individual Sequence: Minimum Description Length}
\label{sec:mdl}

The majority of works in machine learning, especially in deep-learning, assume a distributional perspective 
and rely on the framework of {\em empirical risk minimization} (ERM, \citet{Vapnik1991-gj}) to reason about generalization and evaluation.
In this view an unknown distribution $Q$ is considered the source of the observed 
data $(x, y) \sim Q$ and we strive to learn a model $p(y | x, \theta)$ and parameters
$\theta$ that minimize some loss, for example the log-loss 
$\calL = - \E{x, y \sim Q}{\log p(y | x, \theta)}$.
We have access to a limited number of samples $(x, y) \sim Q$, split them into a training- and a test set, 
and the theory around ERM provides generalization bounds for future data from the same distribution.
This framework relies crucially on a few assumptions; among others, 
that there is a stationary underlying distribution $Q$.

\paragraph{Non-stationary data.} The distributional assumption can become a liability when considering 
non-stationary data. At best, there is some ambiguity 
about the objective a model is supposed to optimize, 
and how to choose a best performing model. 
In general however, the whole notion of a test {\em set} becomes
nonsensical: Sets imply exchangeability, and a 
given problem might not have the invariances that justify exchangability. 
Consider for example the problem of modeling an epidemic outbreak as a function of time and location. 
Selecting random temporal intervals as test set seems naive: interpolating past data between time steps observed in the training data might be much easier than predicting the future. 
Splitting-off test data by location raises a similar dilemma, 
because a model that generalizes well over spatial locations in the past is by no means expected to generalize well on any location into the future.
It seems that choosing an evaluation protocol is a non-trivial and potentially subjective endeavour. 
These ambiguities arise because the concept of a 
test {\em set} implies stationary, exchangable data; 
and in the absence of that, we might be better off not relying on test sets at all.

\paragraph{Compression based inference: Minimum Description Length.} 
Independent of ERM, compression based inference and learning has been widely studied. 
It is based on the fundamental idea that learning and comprehension correspond to compression 
\citep{chaitin2002intelligibility,Hutter:11uiphil}. Given data $\calD{=}(y_t)_1^T$ and a hypothesis 
space $\calM = \{M_1, M_2, \dots \}$, where each hypothesis $M$ corresponds to a parametric probabilistic model 
$p(\calD | \prm, M)$, we aim to identify the model that can compress the data $\calD$ best. 
Considering the close relationship between lossless coding and probability distributions,
this can be achieved by associating a code-length function $L(\calD | M){=}-\log p(\calD | M)$ 
with each hypothesis. 
Additionally, we would have to consider the code-length of the hypothesis $M$ itself to obtain 
the total code length $L(\calD) = L(M) + L(\calD | M)$, which is often ignored if the hypothesis
under consideration are very similar, or are small compared to the data part $L(\calD | M)$.
A vast body of literature argues that models with a shorter description length have
a better chance of generalizing to future data \citep{Wallace2005,Gruenwald:07book,Hutter:11uiphil}.

A crude way to obtain the description length of the data given a {\em parametric} model family is to consider 
$L(\calD|M){=}L_M(\prm){+}L_M(\calD|\prm)$, where $L_M(\prm)$ is the cost of encoding the parameters and $L_M(\calD|\prm){=}{-}\log p(\calD|\prm, M)$ 
is the cost of compressing the data with the parameterized model.
This {\em two-part code} approach might be intuitive but it is sub-optimal.
It is also highly ambiguous because it does not specify how to encode the parameters. 
This crude two-part approach to MDL has been refined in three distinct but closely related ways:
\begin{enumerate}[label=\arabic*),leftmargin=*]
\item The Bayesian marginal likelihood: $L_\text{Bayes}(\calD | M) := -\log \int_\prm p(\calD |  \prm, M) p(\prm) d\prm$ and its variational upper bound $L_\text{ELBO}(\calD | M) := \text{KL}(q(\prm) | p(\prm)) - \E{\prm \sim q}{\log p(\calD | \prm, M)}$. Both  depend on the chosen prior $p(\prm)$ and require access to the posterior distribution over parameters given data; or an approximation $q(\prm)$ thereof. 
\item Normalized Maximum Likelihood (NML): $L_\text{NML}(\calD|M) := -\log\, [p(\calD|\prmmle(\calD), M) \, / Z]$, 
where $\prmmle(\cdot)$ denotes the maximum likelihood estimator $\prmmle(\calD) = \argmax_\prm p(\calD | \prm, M)$ and 
$Z =\int_z p(\calD | \prmmle(z), M) dz]$ is a normalization constant. 
NML normalizes over all possible observable data $z \in \calY^T$, which is often intractable and even undefined for many model families of interest. 
The log-denominator $Z$ is also called the complexity of $M$ and measures how well the model could fit all possible data.
Under NML, a model $M$ achieves a short description length only when it fits the given data well 
(high $\log p(\calD | \prmmle(\calD), M)$) as well as not fit well many different data (low denominator). 
\item  The prequential approach: $L_\text{plugin}(\calD | M) := -\sum_{t=1}^T \log p(y_t | \prmhat(\calD_{<t}), M)$ 
decomposes the description length over datapoints and relies on the choice of a suitable plug-in parameter estimator $\prmhat(\calD_{<t})$. 
It emphasizes that in order to achieve a short codelength, a model $M$ must not only be a good predictor given all training 
data $\calD$, but already given only parts of the data $\calD_{<t}$ for all $t$, i.e.\ be a sample efficient predictor. 
Model complexity and sample efficiency can thus be considered two sides of the same coin. 
\end{enumerate}

Approaches for computing description lengths for some data $\calD$ under a model family 
$p(\calD | \prm , M)$ are called {\em universal codes} when they result in description lengths 
that stay close to the (generally unachievable) maximum likelihood codelength for that family: 
$L_\text{universal}(\calD | M) \le  -\log p(\calD | \prmmle(\calD), M) + \calO(\log T)$. 
$L_\text{NML}(\calD | M)$ is by definition universal because the log-denominator contributes 
a model-architecture dependent gap that is constant in respect to the sequence length.
$L_\text{Bayes}(\calD | M)$ and $L_\text{plugin}(\calD | M)$ are universal for many reasonable choices of the prior $p(\prm)$ and
the estimator $\prmhat(\calD_{<t})$ respectively \citep{Gruenwald:07book}.

Note that none of these make an assumption about the i.i.d-ness of the data $\calD$ or the model $p(\calD | \cdot)$. 
They can all be used to fit a model to an individual sequence $(y_t)_{t=1}^T$ and do not rely on 
held-out sets to evaluate generalization. 
Instead, all three versions of $L(\calD | M)$ contain a build-in form of Occams Razor 
that ensures that models with shorter description length have a better chance to generalize 
to future data. 

Compression based inference is also at the heart of {\em Algorithmic Complexity Theory} (ACT, \citet{Grunwald2004-fm}), 
which provides an alternative foundation to reason about information: 
Where Shannon information is based on distributions, ACT is based on Turing machines and code-lengths. 
Most concepts from Shannons information theory have counterparts in ACT;
e.g: Entropy vs. Kolmogorov complexity, cross-entropy vs. description length, Shannon mutual information vs. 
algorithmic (Kolmogorov) mutual information etc. \citep{Grunwald2004-fm}. 
In ACT, Solomonoff-Induction is the optimal but uncomputable learning and inference machine.
It is closely related to the prequential MDL approch, where we chose the hypothesis space $\calM$ to be the
set of {\em all} computable models (that can be expressed with Turing machines).

\paragraph{The prequential approach}. 

With prequential MDL, learning and model-selection reduce to a specific form of online- or sequential 
learning. The goal is to find a model $M$ that minimizes the description length $L(\calD)$:
\begin{align*}
    L(\calD) 
      &= L(M) + L(\calD | M) \\ 
      &= L(M) - \log p(\calD | M) \\ 
      &= L(M) - \sum_{t=1}^T \log p(y_t | \calD_{<t}, M) \\
\end{align*}
$M$ denotes all modeling choices: The type of model, model-family, model assumptions, 
as well as inference and optimization choices -- everything necessary to make a prediction for 
$y_t$ given $\calD_{<t}$.
For neural networks, it includes the model architecture, parameter initialization,
the optimizer and its hyperparameters, data augmetation strategies, etc. Formally, $M$ is the program
that implements the model and its training procedure on a universal computer such as a Turing machine.
The search space for $M$ is vast, and we can expect that some choices for $M$ lead to significantly 
better models in terms of $\log p(\calD | M) = \sum_{t=1}^T \log p(y_t | \calD_{<t}, M)$ then others. 
However, the fact that we also consider the length $L(M)$ provides a form of Occam's razor and ensures 
a good chance of generalizing to future data:
$L(M)$ grows large if we encode specific information about $\calD$ into $M$ that does not generalize. 

An upper bound for $L(M)$ can be the length of the Python source code of our algorithm, plus the framework 
libraries and all the support code necessary to execute on a general Turing machine -- potentially as a well compressed and self-extracting 
executable. 
In practice, we here compare methods $M_i$ that differ by only a few lines of Python code.
For our purpose the differences in $L(M_i)$ are thus very small compared to $L(\calD | M)$
and we ignore $L(M_i)$. $L(M)$ should however be taken into account when comparing very different kind of learning 
approaches.

\newpage

\section{Priviledged Information Transformer}
\label{app:pit}

We maintain the standard architecture for self-attention transformer blocks \cite{vaswani2017attention}
with parallel feed-foward and attention pathways, however add projections of some privileged information $y_t$
to the keys and values: 

\begin{align*}
    h^{l+1}_t &:= h^l_t + \text{FFW}(\bar{h}^l_t) + \text{Attn}(\bar{h}^l_t) \\ 
    \bar{h}^l_t &:= \text{LayerNorm}(h^l_t) \\
    \text{FFW}(\bar{h}^l_t) &:= W_\text{down} \, \sigma ( W_\text{up} \bar{h}^l_t )  \\
    \text{Attn}(\bar{h}^l_t) &:= \sum_{t'} a_{t, t'} v_{t'} \\
    a_{t, t'} &:= M \odot \text{softmax}_{t'}(q_t \times k_t') \\
    q_{t} &:= W^q \bar{h}^l_t \\
    k_{t} &:= W^k \bar{h}^l_t + \boldsymbol{W^{\bar{k}} y_t} \\
    v_{t} &:= W^v \bar{h}^l_t + \boldsymbol{W^{\bar{v}} y_t} \\ 
\end{align*}

$M$ is a causal attention mask with the diagonal set to 0. Setting the diagonal to 0 ensures that a token at 
time step $t$ can not access the priviledge information in $y_t$ when generating predictions for the current time step.

\section{Piecewise stationary Split-EMNIST experiments}
\label{app:emnist}

We use the convolutional neural network described by \citet{blier2018description} as a feature extractor, 
however double the number of channels in all stages:
The convolutions have  32, 32, 64, 64, 128, 128, 256 and 256 channels per layer respectively 
and max pooling operators after every second layer. 
The convolutions are followed by two fully connected layers of size 256. 
This architecture was used by \citeauthor{blier2018description} to evaluate neural network 
description lengths of MNIST. 
On top, we use 4 transformer blocks and choose a backbone width of 256 units, 
with 4 query-heads with a key-, value- and query- width of 64. 

\subsection{pi-Transformer.}

For comparison we here show the plots from Figures \ref{fig:emnist-plots} and \ref{fig:emnist-detail} in terms
of negative log-loss, instead of error-rates:

\begin{center}
\includegraphics[width=0.45\linewidth]{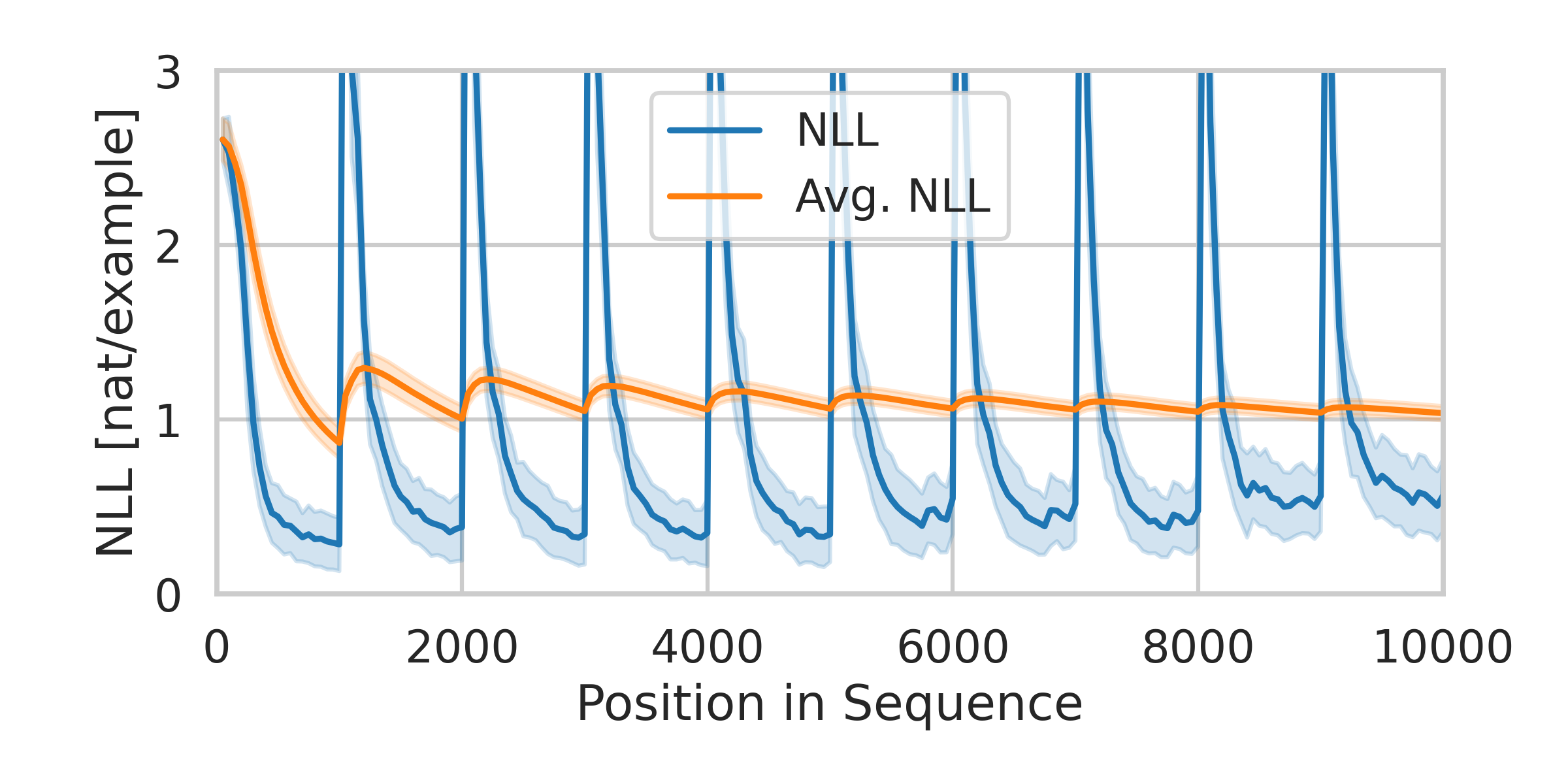}
\hspace{-0.4cm}
\includegraphics[width=0.45\linewidth]{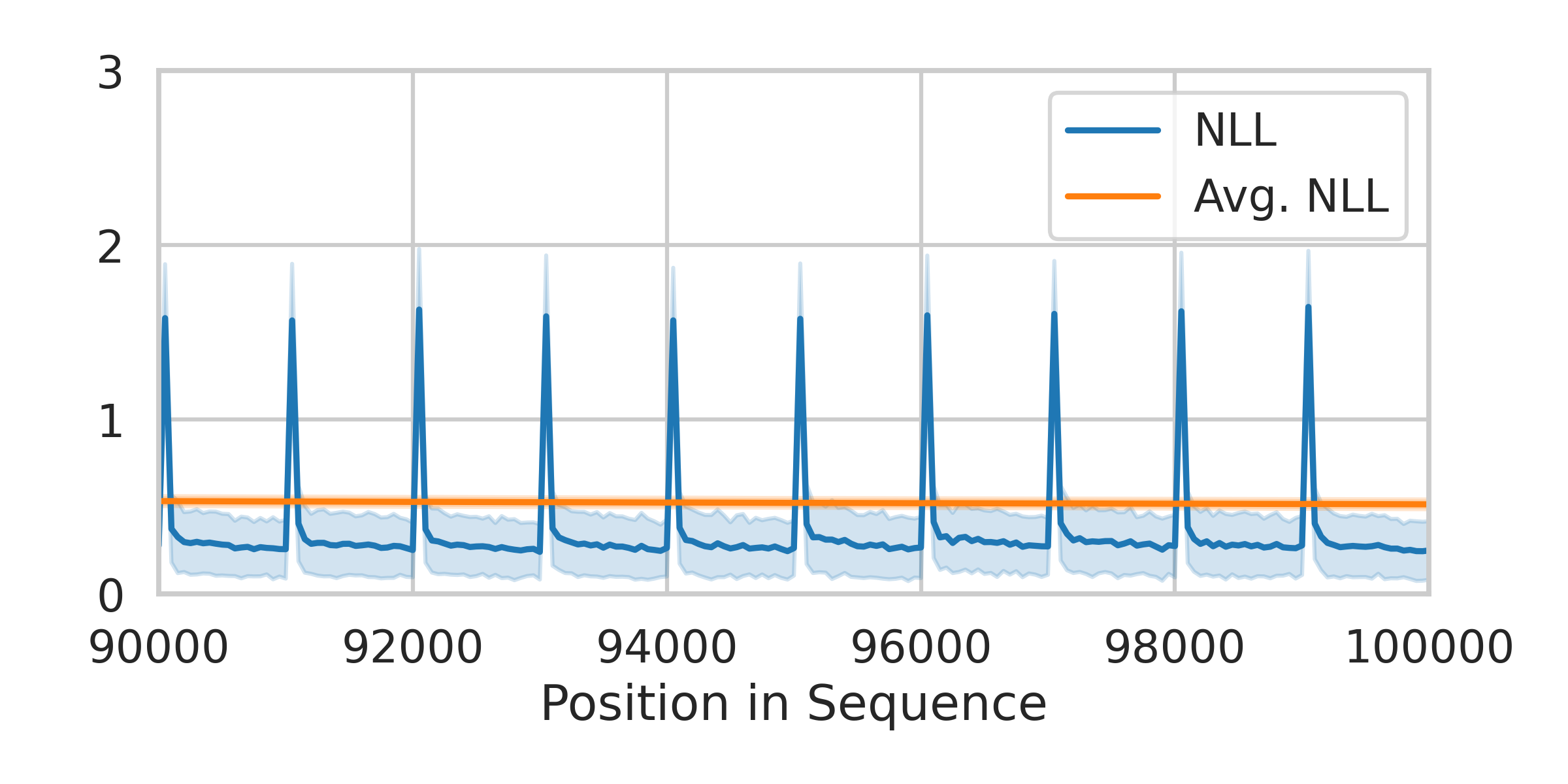}
\end{center}
\vspace{-0.4cm}
Alternative visualization of the Split-EMNIST experiments from Fig. \ref{fig:emnist-plots}:
Instantaneous and averaged next-example log-losses for the first and last 10 tasks of Split-EMNIST: 
Image-to-label mappings are constant within each task, but randomly reassigned at task boundaries    
(averaged over 200 data generating random seeds).
\\
\begin{center}
\includegraphics[width=0.45\linewidth]{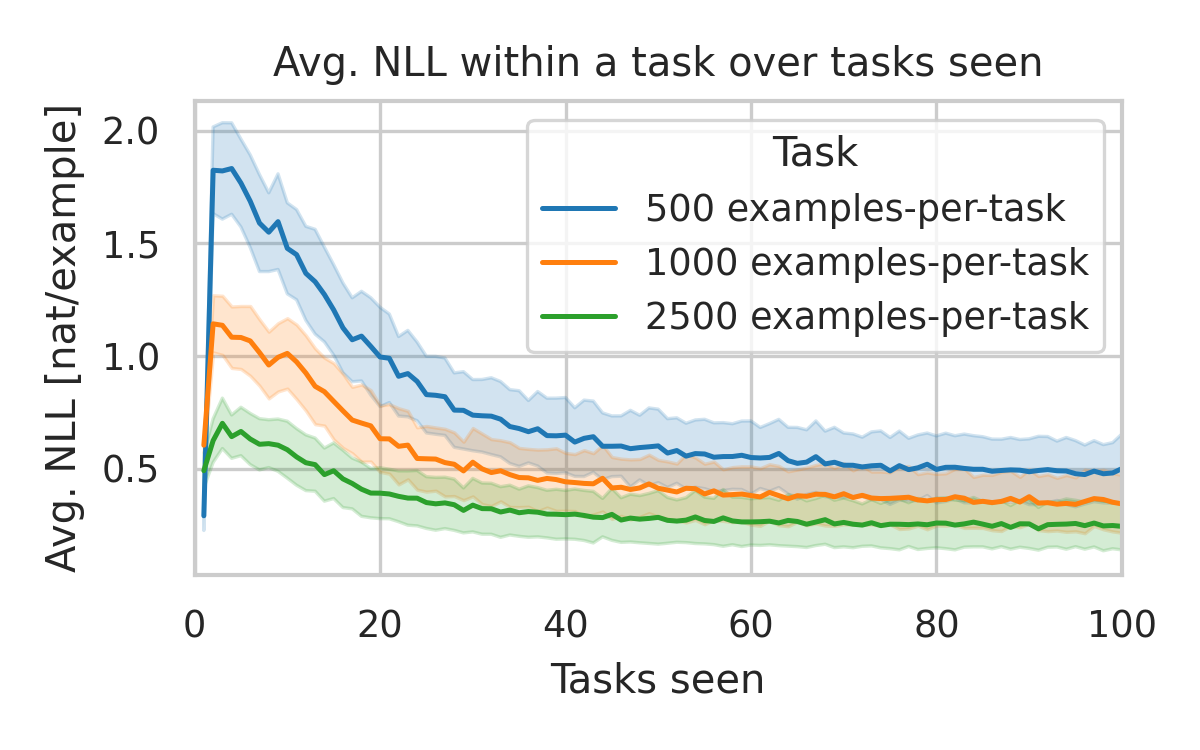}
\includegraphics[width=0.45\linewidth]{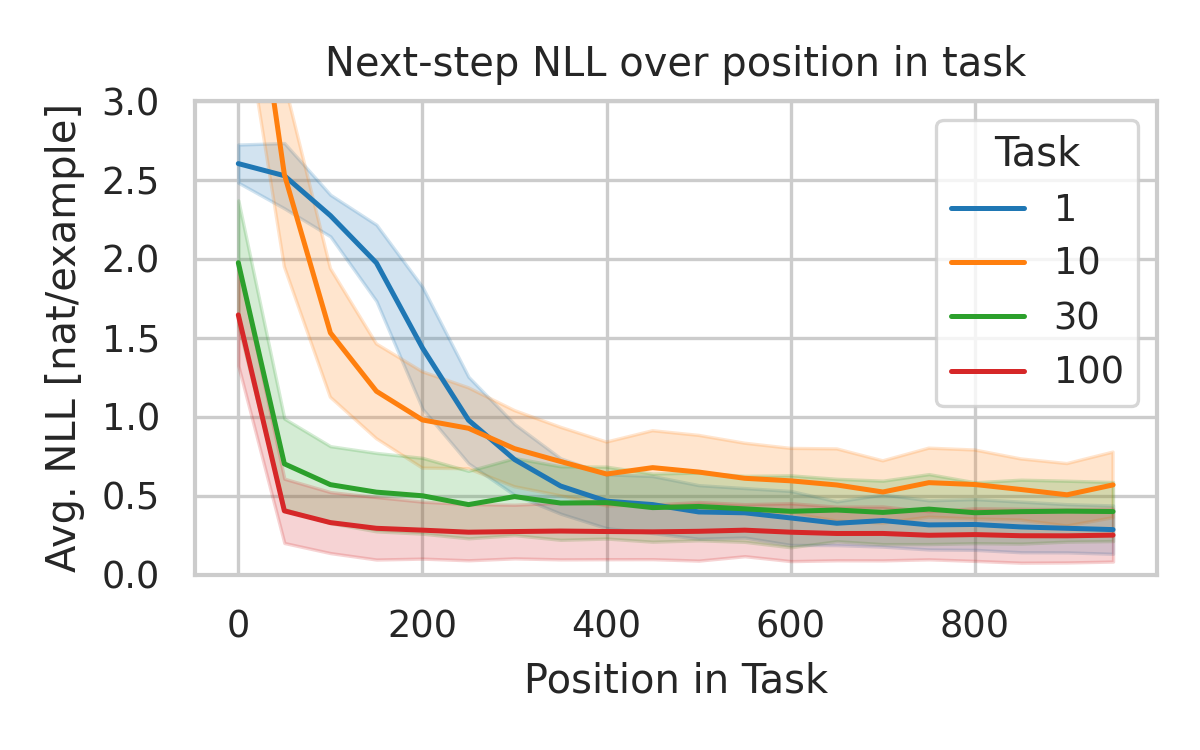}
\end{center}
Alternative visualization of the same experiments:
{\bf Left:} Average performance per task shows strong forward-transfer after struggling during the first 10 to 20 tasks.
{\bf Right:} Detailed look at the within-task performance for the scenario with 1000 examples per task. 
The model is a strong few-shot learner after seeing 30 tasks and further improves until at least task 100

\section{Piecewise stationary Split-CIFAR experiments}
\label{app:cifar}

For CIFAR we use the VGG++ architecture from \citep{bornschein2022sequential} as a feature extractor, 
a VGG-inspired ConvNet with normalization and 10\% dropout added.
On top of that feature extractor, as the temporal model, we use the same 4-block transformer 
architecture as in the Split-EMNIST experiments (Section \ref{sec:emnist_exp}.)

We train with 20 replay streams but otherwise use the same hyper-parameters from the previous section 
(see \ref{app:emnist} for details).
We observe that the results generally resemble those of the Split-EMNIST experiments. 
In particular, the task-averaged predictive performance degrades during the first couple of tasks;
however then rapidly improves and the model soon learns to be an effective few shot learner that makes 
accurate predictions within a few examples after each task switch.

\begin{figure*}[h!]
    \centering
    \includegraphics[width=1.0\linewidth]{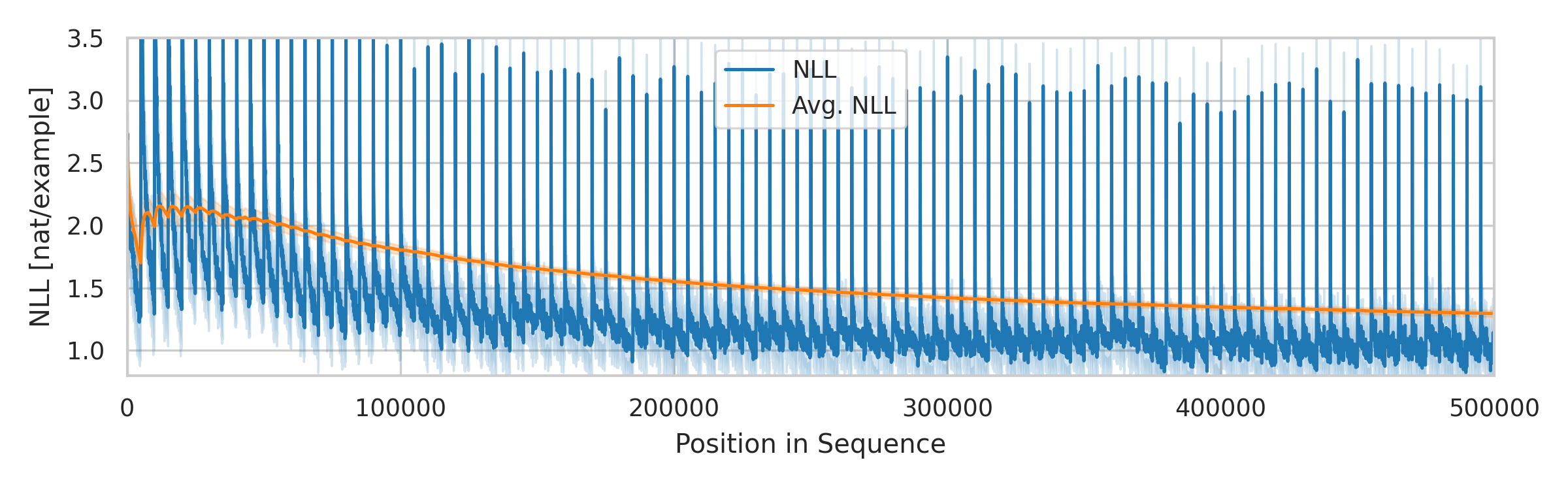}
    \vspace{-0.8cm}
    \caption*{
    Predictive performance on a piecewise-iid CIFAR-100 sequence. The sequence
    consists of 100 tasks with 5000 examples. 
    Each task is a 10-way classification problem, with the 10 classes randomly sourced 
    from the 100 classes of the CIFAR-100 data set. 
    We plot the the instantaneous and averaged next-example prediction performance as a function 
    of examples seen.
    \label{fig:split-cifar-detail}
    }
\end{figure*}

\begin{figure*}[h!]
    \centering
    \includegraphics[width=0.45\linewidth]{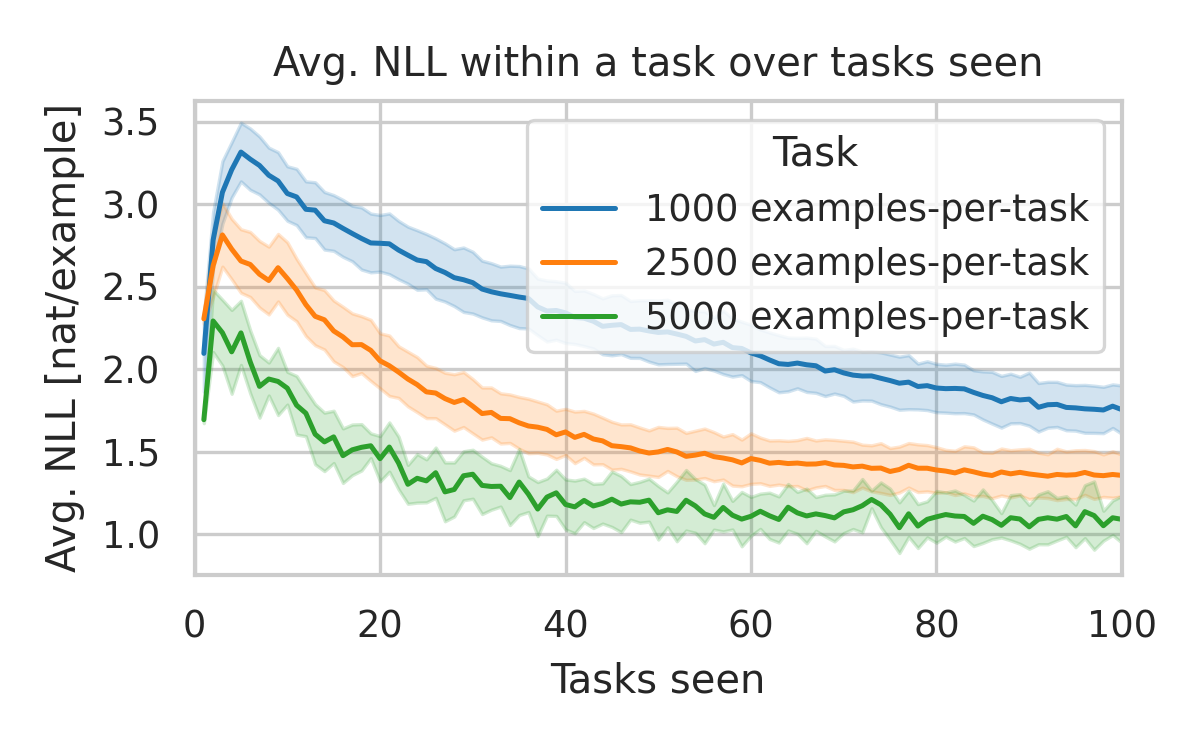}
    \hspace{0.4cm}
    \includegraphics[width=0.45\linewidth]{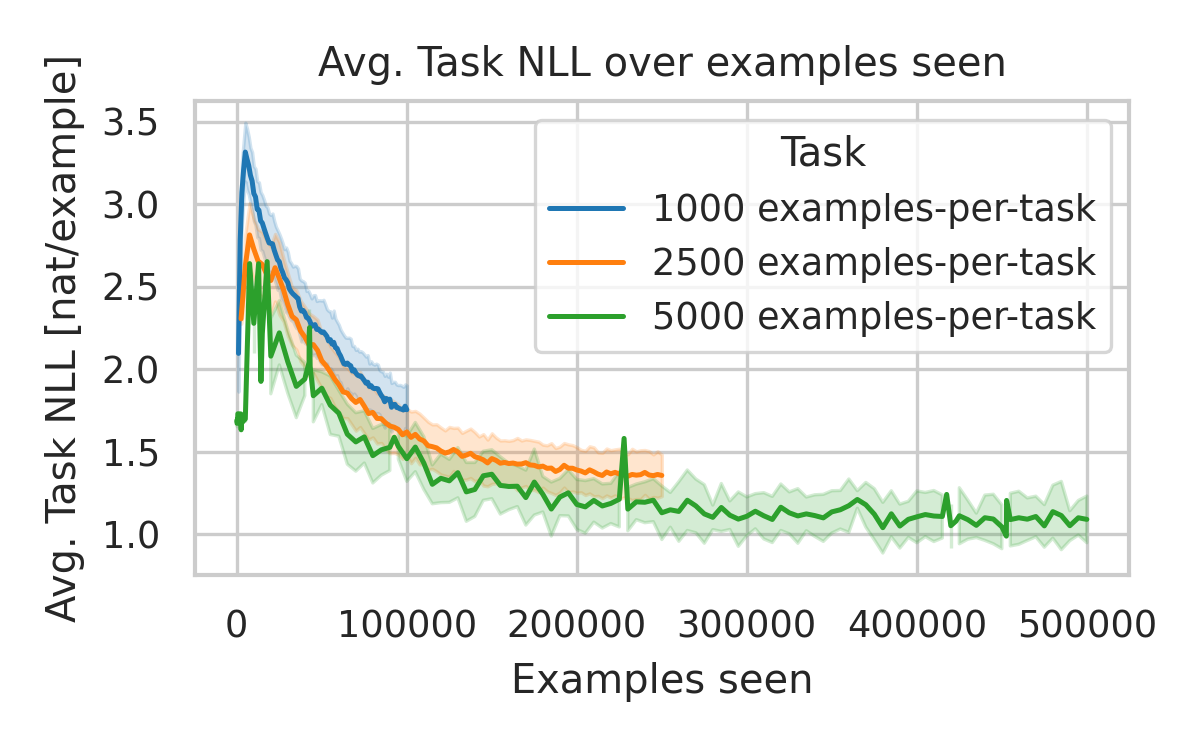}
    \vspace{-0.4cm}
    \caption*{
    Detailed look at predictive performance for piecewise-i.i.d. CIFAR-100:
    We plot the the average per task next-step performance either as a function of 
    tasks seen (left), or as a function of total examples seen (right).
    \label{fig:split-cifar-detail2}
    }
\end{figure*}

\begin{figure*}[h!]
    \centering
    \includegraphics[width=0.45\linewidth]{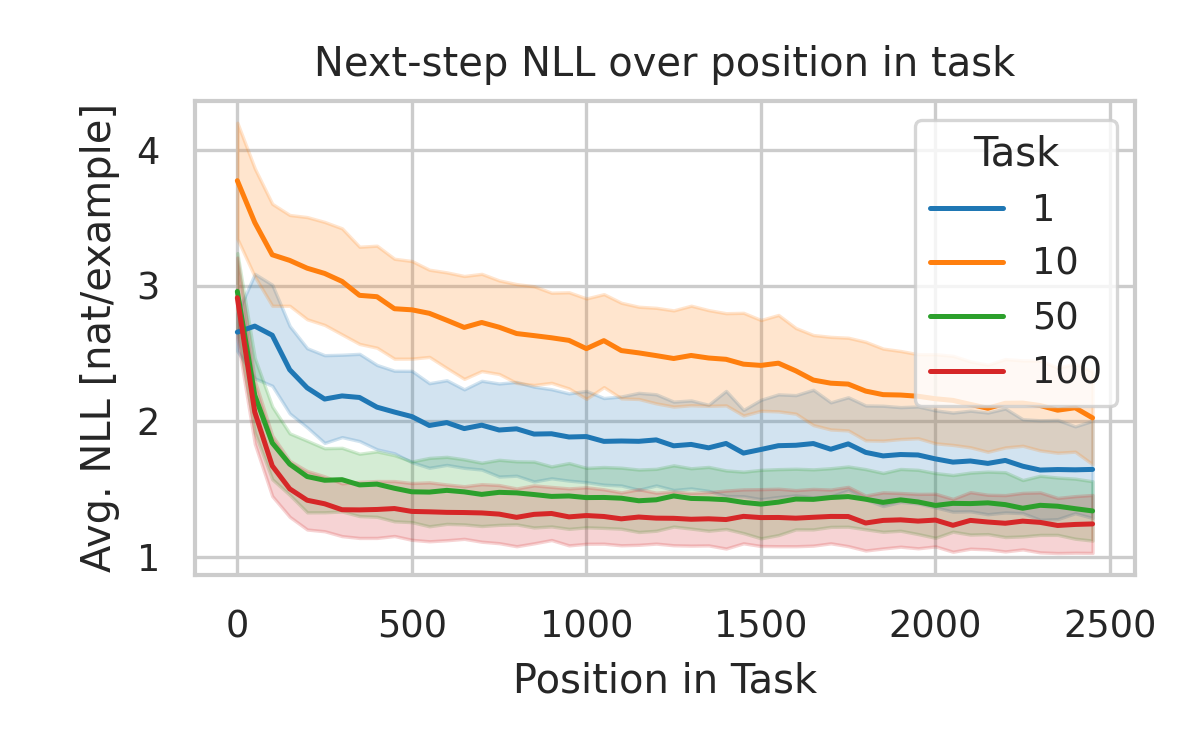}
    \vspace{-0.4cm}
    \caption*{
    Detailed look at the within-task performance for piecewise-iid CIFAR-100 with 2500 examples per task. 
    The model is a strong few-shot learner after seeing 50 tasks and further improves until task 100 
    \label{fig:split-cifar-detail3}
    }
\end{figure*}

\newpage
\section{CLOC with fixed, pre-trained feature extractor}
\label{app:cloc-features}

\subsection{Oracle model}
\label{app:oracle}

\begin{center}
\includegraphics[width=0.49\linewidth]{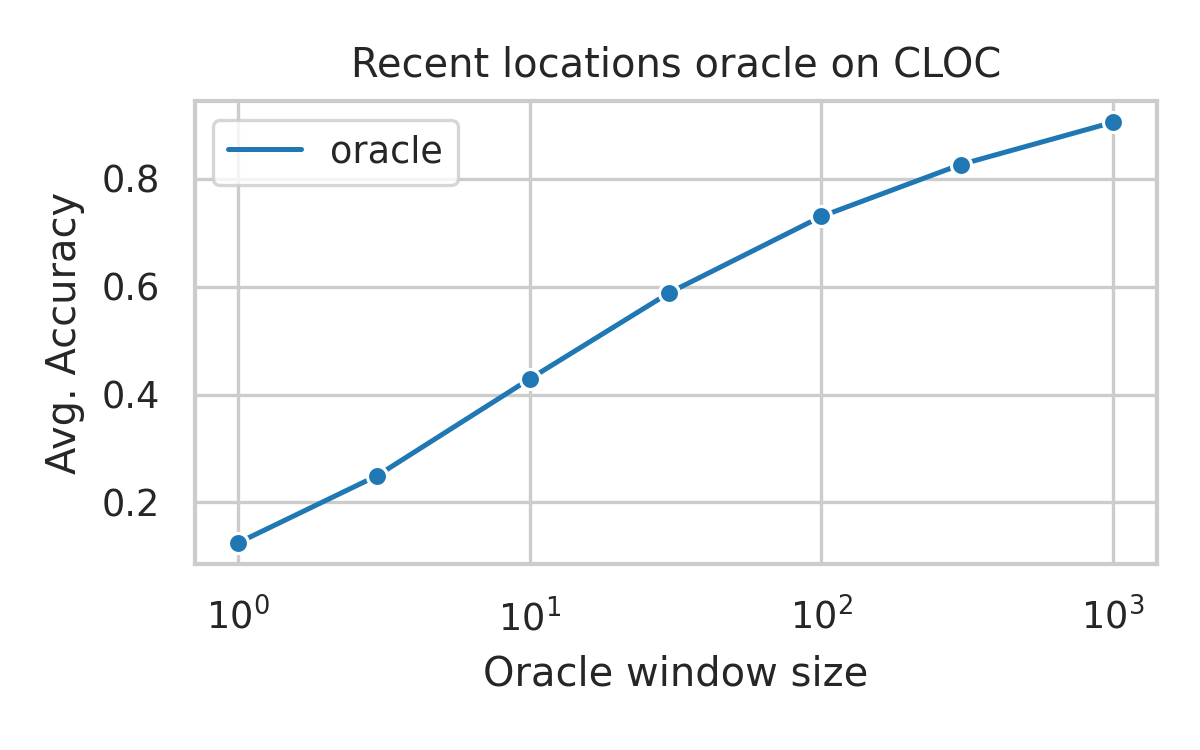}
\end{center}
CLOC sequence: Predictive performance of an oracle model that predicts the correct location 
if and only if that location was observed within a window of the $W$ most recent examples.
The left most data point ($W{=}1 \Leftrightarrow 12\%$ avg. acc.) for example corresponds to a model 
that predicts the correct location if it is the same location as the previous image.
It corresponds to the prediction performance of a model that simply 
predicts $\hat{y_t} = y_{t-1}$.

\subsection{Pi-Transformer with pretrained and fixed Supervised Resnet features}
\label{app:cloc-vitfeatures}

\begin{figure*}[h!]
    \centering
    \includegraphics[width=0.45\linewidth]{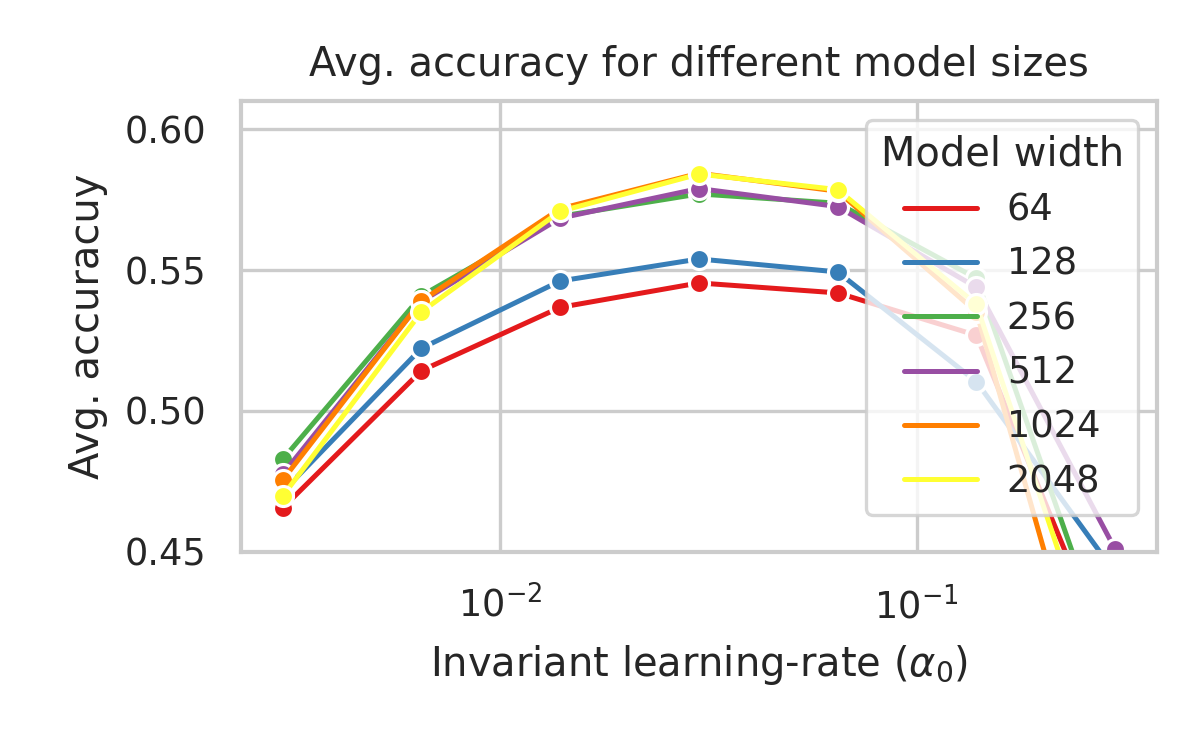}
    \includegraphics[width=0.45\linewidth]{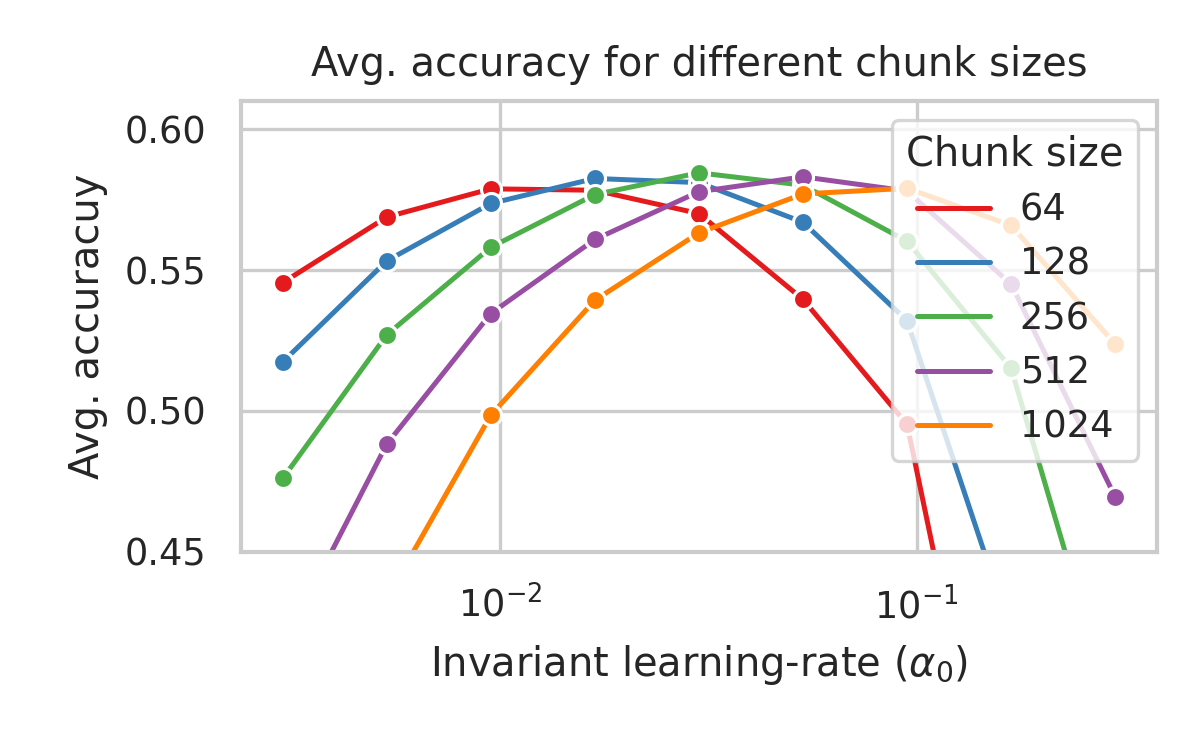}
    \vspace{-0.5cm}
    \caption{
    Pi-Transformer on CLOC with a fixed, pre-trained ResNet-50 feature extractor: 
    {\bf Left)} Larger models are generally better.
    {\bf Right)} Average accuracy is comparably insensitive to the chunk sizes considered here.
    } 
    \label{fig:cloc-sensitivity2}
\end{figure*}

\subsubsection{Pi-Transformer with pretrained and fixed MAE VIT-B features}
\label{app:cloc-vitfeatures}

We run the pi-Transformer with features extracted with the MAE pretrained VIT-B network
from \citet{he2022masked}. 
We show the sensitivity to various hyperparameter changes: 
\\
\begin{center}
\includegraphics[width=0.45\linewidth]{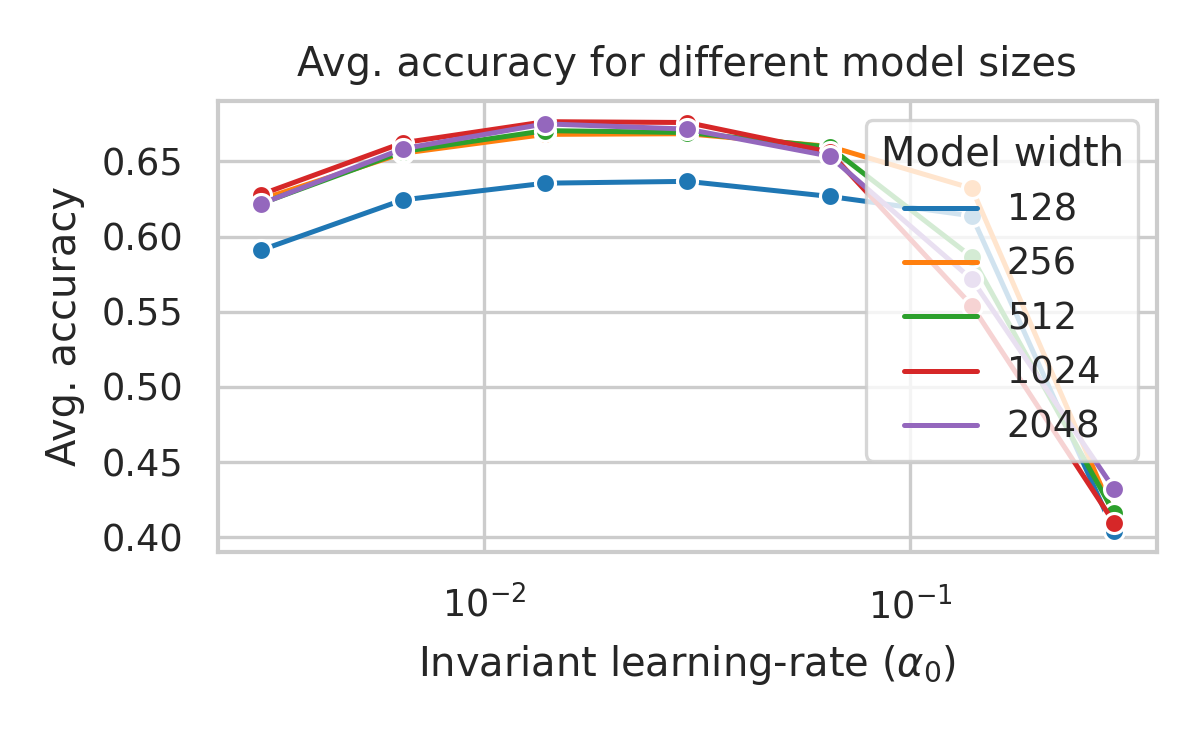}
\includegraphics[width=0.45\linewidth]{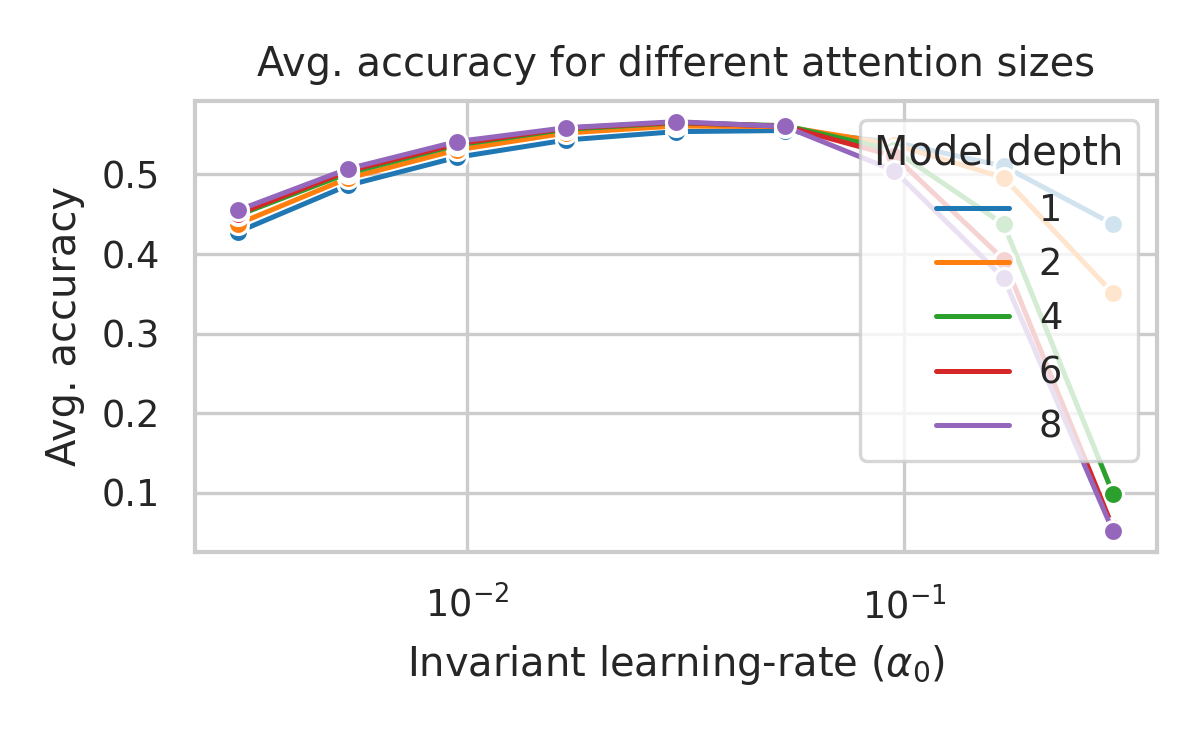} \\
\includegraphics[width=0.45\linewidth]{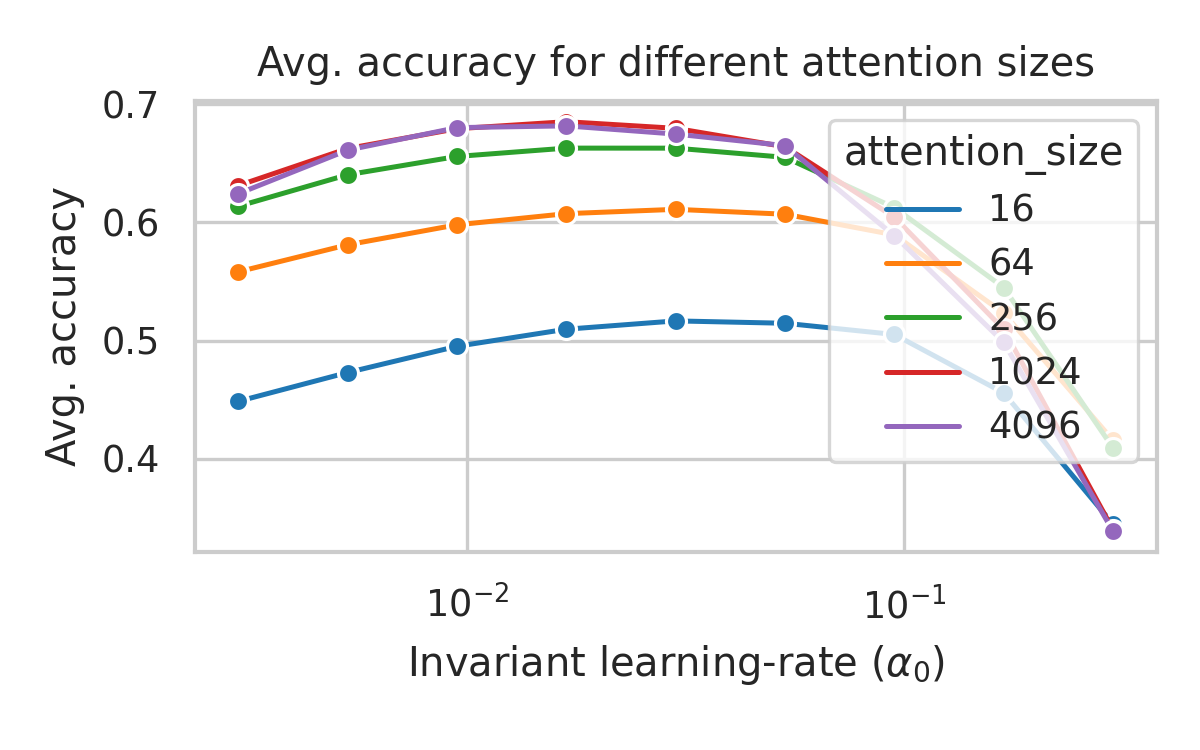}
\includegraphics[width=0.45\linewidth]{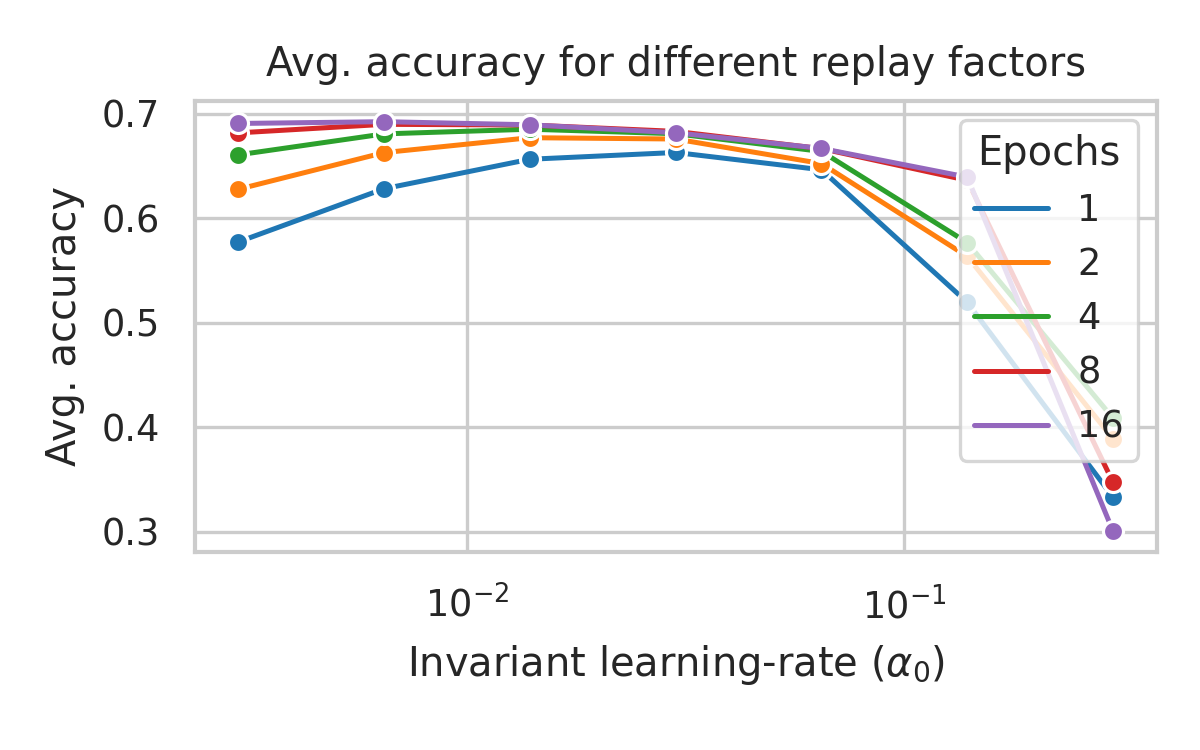}
\end{center}

\subsection{2-token approach with pretrained and fixed Resnet features}
\label{app:cloc-features-2t}

We online train a Transformer with the 2-token approach on a with features extracted 
with a Resnet-50 trained on supervised ImageNet 1k.
The transformer has a backbone-width of $D=1024$ and is 8 blocks deep, which 
is twice as deep as the 4 block deep architecture we usually use for pi-Transformers 
on pre-extracted CLOC features.
We show the sensitivity to various hyperparameter changes: \\
\begin{center}
\includegraphics[width=0.45\linewidth]{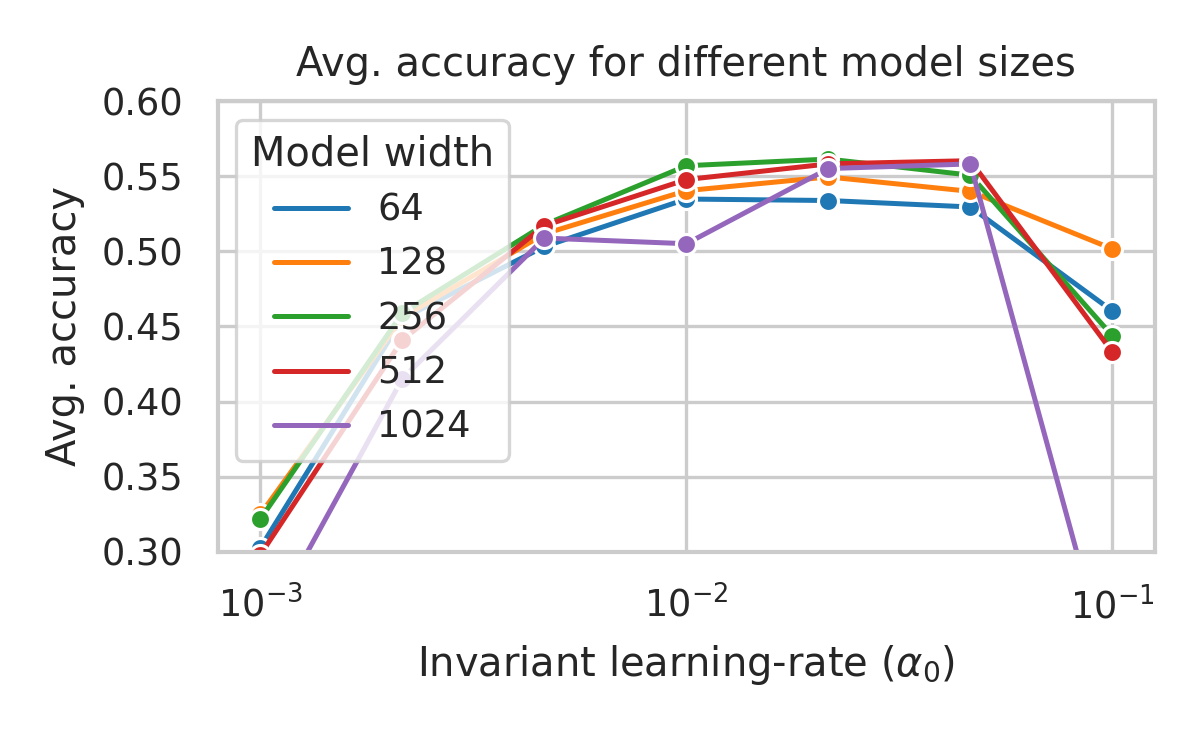}
\includegraphics[width=0.45\linewidth]{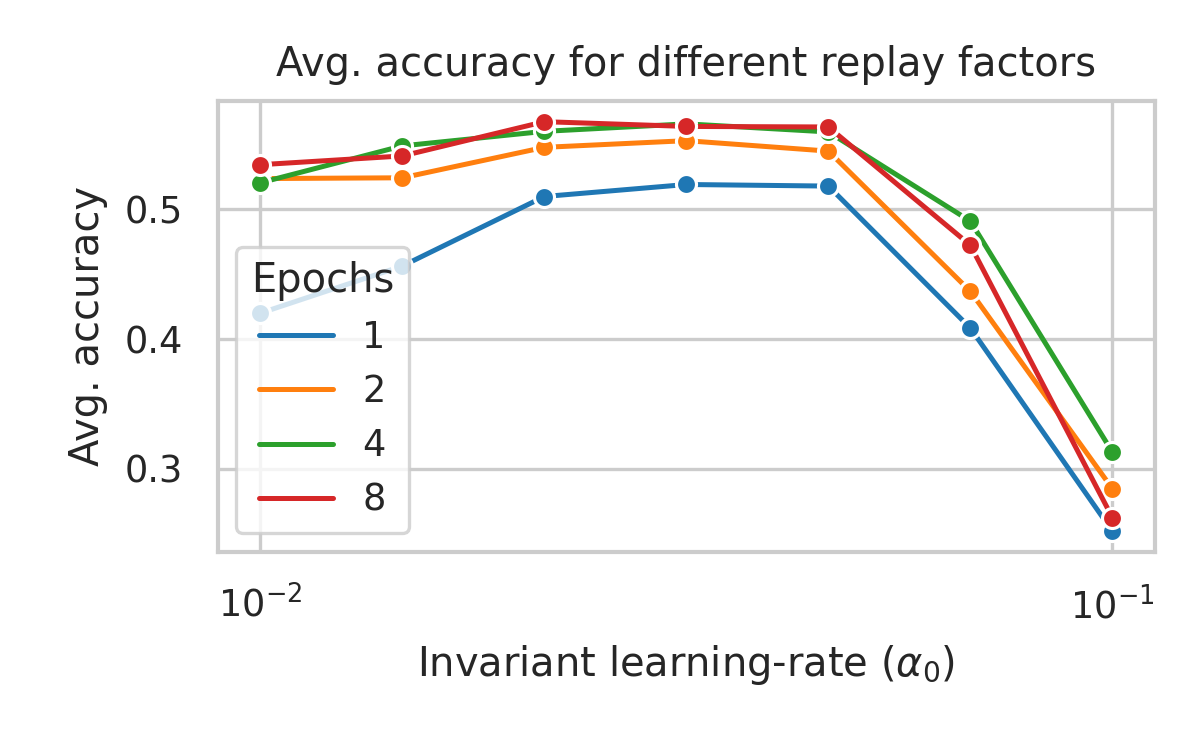} \\
\includegraphics[width=0.45\linewidth]{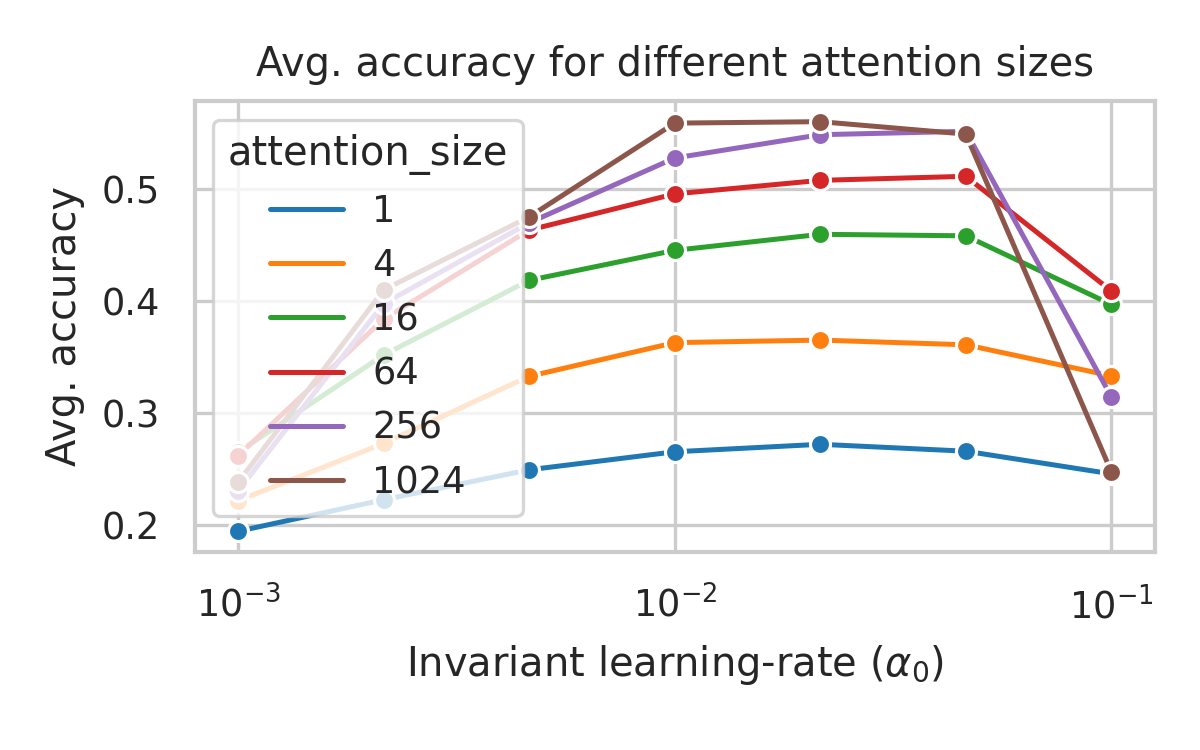}
\includegraphics[width=0.45\linewidth]{figures/cloc-features-inc.png}
\end{center}

\subsection{Hyper-parameter cube for Pareto fronts with pretrained and fixed feature extractors}
\label{app:cloc-features-hypercube}

We use the same hyper parameter search space for all experiments:

\begin{center}
\begin{tabular}{rl}
    \toprule
    Hyper Parameter & Range \\
    \midrule
     Model Width ($D$) & $[16, \cdots, 4096)$ \\
     Model Depth & $[1, \cdots, 12]$ \\
     Num Streams & $[1, \cdots, 16]$ \\
     Attention window ($C$) & $[128, \cdots, 2048]$ \\
     Learning Rate ($\alpha_0$) & [1e-2, $\cdots$, 1e-2] \\ 
     AdamW Weight Decay  & [1e-2, $\cdots$, 3e-2] \\
     \bottomrule
\end{tabular}
\end{center}

\section{CLOC with ConvNet feature extractor from scratch}
\label{app:cloc-resnet}

\subsection{Description of the feature extractor}

We use the ConvNext tiny architecture \citep{liu2022convnet} as a feature extractor and feed the top-level activations as inputs to the temporal transformer.  
We scale input images to a resolution of $160 \times 160 \times 3$ channels and run them 
initially through a convolution with stride 4 and 96 channels. 
These features are then processed by 4 successive stages with 96, 192, 384, 768 channels respectively. 
Between each stage we downscale the spatial resolution by a factor of 2 and apply LayerNorm.. 
The stages have 3, 3, 9, 3 blocks each. The pseudo-code for each block reads:

\begin{verbatim}
def forward(input)
    x = DepthwiseConv2D(1, kernel_shape=7)(input)
    x = LayerNorm(axis=-1, create_scale=True, create_offset=True)(out)
    x = self.activation(
    x = Linear(4 * channels)(out)
    x = Linear(channels)(out)
    return input + x * get_parameter('skip_gain', x.shape[-1])
\end{verbatim}

\subsection{Hyper-parameters for Pi-Transformer experiments}

\begin{center}
\begin{tabular}{rl}
    \toprule
    Hyper Parameter & Value \\
    \midrule
    Image Size & $160 \times 160$ \\ 
    Feature Extractor Learning Rate & 3e-4 \\ 
    Transformer Width ($D$) & 1024 \\
    Model Depth & 2 \\
    Num Streams & 8 \\
    Chunk Size ($S$) & 256 \\
    Attention window ($C$) & 2048 \\
    Transformer Learning Rate ($\alpha_0$) & 1e-2 \\ 
     \bottomrule
\end{tabular}
\end{center}

\subsection{Hyper-parameters for online Transformers with 2-token approach}

\begin{center}
\begin{tabular}{rl}
    \toprule
    Hyper Parameter & Value \\
    \midrule
    Image Size & $160 \times 160$ \\ 
    Feature Extractor Learning Rate & 3e-4 \\ 
    Transformer Width ($D$) & 1024 \\
    Model Depth & 10 \\
    Num Streams & 8 \\
    Chunk Size ($S$) & 256 \\
    Attention window ($C$) & 2048 \\
    Transformer Learning Rate ($\alpha_0$) & 1e-2 \\ 
     \bottomrule
\end{tabular}
\end{center}

\begin{align*}
    p(y_{t} | x_{t}, \calD_{<t}, \theta_{t-1}) \\
\end{align*}

\end{document}